\documentclass{article}

\PassOptionsToPackage{numbers, square, sort}{natbib}

\PassOptionsToPackage{colorlinks=true,linkcolor=linkpink,citecolor=linkpink,urlcolor=linkpink}{hyperref}

    \usepackage[preprint]{neurips_2024}

\usepackage[utf8]{inputenc} % allow utf-8 input
\usepackage[T1]{fontenc}    % use 8-bit T1 fonts
\usepackage{hyperref}       % hyperlinks
\usepackage{url}            % simple URL typesetting
\usepackage{booktabs}       % professional-quality tables
\usepackage{amsfonts}       % blackboard math symbols
\usepackage{nicefrac}       % compact symbols for 1/2, etc.
\usepackage{microtype}      % microtypography
\usepackage{xcolor}         % colors

\usepackage{microtype}
\usepackage{graphicx}
\usepackage{tabularx}
\usepackage{booktabs} % for professional tables

\usepackage{hyperref}

\usepackage{multirow}
\usepackage{multicol}
\usepackage{makecell} 
\usepackage{xcolor}
\usepackage{colortbl}
\usepackage{float}
\usepackage{parskip}
\usepackage{graphicx}
\usepackage{pifont}
\usepackage{enumitem}

\usepackage{subfig}
\usepackage{algorithm}
\usepackage{algpseudocode}
\usepackage{wrapfig}
\usepackage{lipsum}

\usepackage{amsmath}
\usepackage{amssymb}
\usepackage{mathtools}
\usepackage{amsthm}

\usepackage[capitalize,noabbrev]{cleveref}

\usepackage{amsmath}
\usepackage{amssymb}
\usepackage{mathtools}
\usepackage{amsthm}
\usepackage{xparse}
\usepackage{color}
\usepackage{xcolor}
\definecolor{linkpinkix}{HTML}{EA335A} % was blue; now the attached pink

\definecolor{linkpink}{HTML}{EA335A}

\definecolor{lime}{HTML}{A6CE39}

\usepackage[colorlinks=true,linkcolor=linkpink,citecolor=linkpink]{hyperref}

\theoremstyle{plain}

\theoremstyle{definition}

\theoremstyle{remark}

\usepackage[textsize=tiny]{todonotes}

\usepackage{xspace}

\newcommand{\gx}[1]{\tilde{\boldsymbol{x}}_{#1}}
\newcommand{\lxx}[2]{\boldsymbol{x}_{#2}^{(#1)}}
\newcommand{\lx}[3]{\boldsymbol{x}_{#2, #3}^{(#1)}}
\newcommand{\optx}{\boldsymbol{x}_*}
\NewDocumentCommand\x{ggg}{%
    \IfNoValueTF{#1}{\optx}{\IfNoValueTF{#2}{\gx{#1}}{\IfNoValueTF{#3}{\lxx{#1}{#2}}{\lx{#1}{#2}{#3}}}}
}

\newcommand{\globalmask}{\mathcal{M}}
\newcommand{\localmask}[1]{\mathcal{M}^{(#1)}}
\newcommand{\localmaskwithelement}[2]{\mathcal{M}^{(#1)}_{#2}}
\NewDocumentCommand\mask{gg}{%
    \IfNoValueTF{#1}{\globalmask}{\IfNoValueTF{#2}{\localmask{#1}}{\localmaskwithelement{#1}{#2}}}
}

\renewcommand{\eqref}[1]{Equation (\ref{#1})}

\newcommand{\bracket}[1]{\left(#1\right)}

\newcommand{\stogradone}[1]{\tilde{g}_{#1}}
\newcommand{\stogradthree}[3]{g^{(#1)}_{#2, #3}}
\NewDocumentCommand\g{ggg}{%
    \IfNoValueTF{#2}{\stogradone{#1}}{\stogradthree{#1}{#2}{#3}}
}

\usepackage{amsmath}

\DeclareMathOperator*{\argmin}{arg\,min}

\newcommand{\gradone}[1]{\nabla F\left(\gx{#1}\right)}
\newcommand{\gradtwo}[2]{\nabla F_{#1}\left(\gx{#2}\right)}
\newcommand{\gradthree}[3]{\nabla F_{#1}\left(\x{#1}{#2}{#3} \odot \mask{#1}{#2}\right) \odot \mask{#1}{#2}}
\NewDocumentCommand\grad{mgg}{%
    \IfNoValueTF{#2}{\gradone{#1}}{\IfNoValueTF{#3}{\gradtwo{#1}{#2}}{\gradthree{#1}{#2}{#3}}}
}

\newcommand{\partialmark}{\texttildelow}

\newcommand{\na}{\textemdash}

\newcommand{\newmaskgradtwo}[2]{\nabla_{\x{#2}} F_{#1}\bracket{\x{#2} \odot \mask{#1}{#2} \bracket{\x{#2}}}}
\newcommand{\newmaskgradthree}[3]{\nabla_{\x{#1}{#2}{#3}} F_{#1}\bracket{\x{#1}{#2}{#3} \odot \mask{#1}{#2} \bracket{\x{#1}{#2}{#3}}}}
\newcommand{\newmaskgradfour}[4]{\nabla_{\x{#1}{#2}{#3}} F_{#1}^{[\gamma_{#4-1}':\gamma_{#4}']} \bracket{\x{#1}{#2}{#3} \odot \mask{#1}{#2} \bracket{\x{#1}{#2}{#3}}}}
\NewDocumentCommand\newmaskgrad{mggg}{%
    \IfNoValueTF{#2}{\gradone{#1}}{\IfNoValueTF{#3}{\newmaskgradtwo{#1}{#2}}{\IfNoValueTF{#4}{\newmaskgradthree{#1}{#2}{#3}}{\newmaskgradfour{#1}{#2}{#3}{#4}}}}
}

\newcommand{\globaloutergradgammacons}[2]{\nabla F^{[\gamma_{#1-1}': \gamma_{#1}']}\bracket{\x{#2}}}

\newcommand{\globaloutergradgammaconsthree}[3]{\nabla F_{#3}^{[\gamma_{#1-1}': \gamma_{#1}']}\bracket{\x{#2}}}
\NewDocumentCommand\gggamma{mgg}{%
    \IfNoValueTF{#2}{\gradone{#1}}{\IfNoValueTF{#3}{\globaloutergradgammacons{#1}{#2}}{\globaloutergradgammaconsthree{#1}{#2}{#3}}}
}

\renewcommand{\cite}{\citep}

\title{\texttt{UnifiedFL}: A Dynamic Unified Learning Framework for Equitable Federation}

\author{%
  Furkan Pala, Islem Rekik\\
  BASIRA Lab, Imperial-X (I-X) and Department of Computing\\
  Imperial College London, London, United Kingdom\\
  \texttt{\{f.pala23, i.rekik\}@imperial.ac.edu} 
}

\begin{document}

\maketitle

\begin{abstract}
    
Federated learning (FL) has emerged as a key paradigm for collaborative model training across multiple clients without sharing raw data, enabling privacy-preserving applications in domains such as radiology and pathology. However, works on collaborative model training across clients with fundamentally different neural architectures and non-identically distributed datasets remain largely scarce. Besides, existing FL frameworks face several limitations. First, despite claiming to support architectural heterogeneity, most recent FL methods only tolerate variants within a single model family---such as shallower, deeper, or wider CNNs---thereby still presuming a shared global architecture and failing to accommodate federations in which clients deploy fundamentally different network types (e.g., CNNs, GNNs, MLPs). Second, existing approaches often address only the statistical heterogeneity of datasets across clients yet overlook the domain-fracture problem, where each client’s training data stem from distributions that differ markedly from those faced at testing time, an oversight that severely undermines the generalizability of every client model. More importantly, when clients use different model architectures and have differently distributed data---and the test data differ yet again---current methods cause each client’s model to perform poorly. 
To address such challenges, we propose \texttt{UnifiedFL}, a dynamic \emph{unified} federated learning framework that represents heterogeneous local networks as nodes and edges in a directed \emph{model-graph}, whose weights and biases are optimized by a single, shared graph neural network (GNN). Our three core contributions lie in (i) parameterizing all local architectures through a common GNN, ensuring that incompatible tensors are never transmitted; (ii) introducing a distance-driven clustering mechanism based on Euclidean distances between clients’ GNN parameters to dynamically group hospitals following similar optimization trajectories; and (iii) designing a two-tier aggregation policy that synchronizes frequently within clusters while sparsely across clusters to balance convergence and diversity. Our comprehensive experiments on four MedMNIST disease classification benchmarks and the Hippocampus segmentation task from the Medical Image Segmentation Decathlon demonstrate the outperformance of \texttt{UnifiedFL} over strong FL baselines on both classification and segmentation metrics. \texttt{UnifiedFL} presents the first framework to unify heterogeneous model training in FL via a shared model-graph representation. Our Python \texttt{UnifiedFL} code, benchmarks, and evaluation datasets are available at \url{https://github.com/basiralab/UnifiedFL}.
\end{abstract}

\section{Introduction}\label{sec:intro}

Medical artificial intelligence (AI) has become an integral component of clinical decision-making pipelines, offering unprecedented capabilities in medical image analysis for disease screening, prognosis, and therapeutic planning \cite{Pan_PGMLIF_MICCAI2024, wang2019artificial,rajpurkar2022ai}. These gains are most evident in data-hungry fields such as radiology and pathology, where advanced machine learning algorithms, including deep neural networks (DNNs), can extract patterns from high-dimensional imaging data to aid clinicians by improving diagnostic accuracy, forecasting disease progression, and informing treatment decisions. However, building high-performing and clinically reliable models often necessitates pooling extensive, diverse datasets from multiple hospitals. Such data-sharing remains fraught with privacy, regulatory, and logistical challenges, especially under stringent rules like HIPAA and GDPR \cite{rieke2020future, kairouz2021advances}. To overcome the hurdles of data centralization, \emph{federated learning} (FL) has emerged as a transformative approach, allowing each hospital (or ``client'') to train models locally while sharing only model updates with a central server, then aggregating model and broadcasting model parameters back to clients without sharing data. \cite{pmlr-v54-mcmahan17a,xu2021federated,Li_SiFT_MICCAI2024,guan2024federated,adnan2022federated}. This privacy-preserving paradigm has proven particularly valuable in healthcare, where sensitive patient data cannot be moved freely beyond clinical boundaries. Nonetheless, FL in medical imaging must address additional complexities. The key challenges are (1) heterogeneity of the local computational environment and network architecture, and (2) non-identical, non-independent (\emph{non-IID}) data distributions driven by factors like demographic discrepancies across hospitals.

In medical imaging, different neural network architectures tend to perform better on different types of tasks, depending on the nature of the data and the clinical objective. Convolutional neural networks (CNNs) are preferred for tasks that rely heavily on spatial context, while multi-layer perceptrons (MLPs) or Transformers may be employed for tabular data, time-series records, or high-resolution scans with complex texture features. As a result, healthcare local clients may develop or adopt models with divergent layer types, widths, depths, or input modalities \cite{adnan2022federated}. Conventional FL algorithms, such as FedAvg \cite{pmlr-v54-mcmahan17a} assumes every client runs an identical network, so its server can merge updates position by position. When confronted with \emph{fully heterogeneous models}, direct aggregation fails due to mismatched weight shapes and inconsistent layer definitions \cite{pmlr-v54-mcmahan17a,adnan2022federated}. Although certain approaches alleviate partial heterogeneity by limiting variations to specific layers (e.g., an extra personalization layer or different output head), the complete unification of \emph{entirely} different network architectures remains a significant challenge in federated healthcare applications.

Beyond architectural diversity, medical data typically exhibit strong domain shifts across local clients, often due to variations in scanner types, image acquisition protocols, and patient populations \cite{rieke2020future,kairouz2021advances}. In practice, the frequency of disease classes can differ drastically from one hospital to another, or image quality may vary based on local hardware. Consequently, models trained under a strict assumption of IID data distribution may perform poorly in real-world federated scenarios, demonstrating low robustness and generalizability \cite{xu2021federated,Li_SiFT_MICCAI2024}. Techniques such as cluster-based FL \cite{sattler2020clustered} or personalization \cite{adnan2022federated,sattler2020clustered} have been proposed to mitigate these effects, yet many still rely on a relatively consistent global model architecture, preventing straightforward application in a federated setting involving diverse model architectures.

A promising line of research addresses both \emph{architectural} and \emph{statistical} diversity by embedding disparate models into a shared parameter space, often via graph-based representations. Instead of matching weights index-by-index, each network is converted into a graph structure, where nodes and edges represent biases and weights (or filters), respectively. A graph neural network (GNN) is then employed to align these `model-graphs,'' enabling a single set of GNN parameters to drive updates across otherwise incompatible architectures \cite{adnan2022federated}. In conjunction with iterative client-server exchanges, this approach opens the door for a \emph{truly model-agnostic FL} pipeline. Yet, when facing non-IID data, the frequency and manner of communication become crucial: frequent interactions among distinct topologies can lead to parameter interference while too little communication hampers knowledge sharing. To regulate aggregation, one strategy is to group clients with similar network topologies into clusters and reduce cross-cluster interactions, assuming that similarly structured models benefit from more frequent parameter exchange \cite{adnan2022federated,sattler2020clustered}. However, relying on a static and \emph{a priori} defined grouping criteria, e.g., purely topological features such as node degrees or network depth may not be always suitable. Once these clusters are formed at initialization, they remain fixed throughout training. Realistically, however, medical local clients may continuously update or adapt their local model designs, or discover that their learned weights drift significantly as new cases and imaging protocols are introduced \cite{kairouz2021advances,sattler2020clustered}. A static clustering scheme thus risks suboptimal groupings, leading to diminished collaboration among clients who become more aligned over time, or forced interactions among those who diverge as training evolves.

In this paper, we present \texttt{UnifiedFL}: a dynamic unified federated learning framework for equitable medical imaging, which aims to address fully heterogeneous architectures, domain shifts, and the evolving nature of local models in federated medical imaging. Our contributions are listed below:

\begin{itemize}
    \item \textbf{Unified learning:} We unify each neural network across clients (be it CNN or MLP) by transforming them into a model-graph representation so that local updates can be performed under a common GNN-based parameterization. Thus, clients and server communicate only the GNN parameters regardless of the heterogeneity of the neural networks in clients, making our federation \emph{truly architecture-agnostic}.
    
    \item \textbf{Dynamic clustering:} After each communication round every client sends the latest values of its shared GNN parameters to the server. The server measures pair-wise distances between these vectors, groups nearby clients into clusters, and updates the groups at regular intervals. Hospitals whose optimization paths converge are therefore synchronized often, whereas those that drift apart exchange updates less frequently, preventing harmful interference.

    \item \textbf{Enhanced robustness to non-IID data:} Our approach inherently handles the statistical heterogeneity of client data distributions so that each client can train its local model on its own data. This allows each client to select the most appropriate architecture for the statistical distribution of their data. Next, our dynamic clustering mechanism mitigates domain shift among clients’ training datasets by grouping together clients with similar training dynamics, enabling personalized yet collaborative learning within each cluster.
    
    % To mimic real-world non-IID conditions, we cluster similar data samples and assign each cluster to a different client, thereby introducing a domain shift across clients. This setup simulates realistic distributional disparities, as encountered in federated healthcare settings where each hospital holds data from distinct populations.
    
    % \item \textbf{Focus on equitable healthcare:} The proposed method is designed to facilitate fair collaborations across institutions of differing sizes, resources, and patient demographics. A dynamic collaboration mechanism mitigates the risk of excluding resource-limited sites or overshadowing minority populations, leading to a more equitable AI ecosystem.
\end{itemize}

\section{Related work}
\label{sec:related}

FL in medical imaging has recently expanded beyond the classical assumption that every site trains an identical network on similarly distributed data.  Two research streams tackle the ensuing challenges: \textit{(i) heterogeneous-model aggregation}, which aims to merge updates from clients that run \emph{different} architectures, and \textit{(ii) knowledge-distillation frameworks}, which sidestep parameter alignment by training lightweight models called student networks under the supervision of a larger model called teacher network.  We review both streams, emphasizing the gaps that motivate \texttt{UnifiedFL}.

\paragraph*{Heterogeneous federated learning}
\textit{HeteroFL} \cite{diao2021heteroflcomputationcommunicationefficient} prunes each convolutional layer width-wise so that the convolutional filters align across clients before averaging, but pruning discards low-level features and locks the pruning ratio at design time.  \textit{InclusiveFL} \cite{inclusiveFL} attaches a shallow “student’’ network to smaller devices and averages overlapping layers with deeper “teacher’’ models; yet the depth split is static, and mismatch in layer types (e.g.\ depthwise vs.\ standard convolutions) is still disallowed.  \textit{ScaleFL} \cite{scaleFL} searches width–depth pairs and adds early-exit heads, improving resource adaptivity, though the search space must be tuned for each architecture and does not adapt after deployment.  Parameter-efficient approaches freeze the layers and aggregate small branches: \textit{pFedLoRA} \cite{yi2024pfedloramodelheterogeneouspersonalizedfederated} uses low-rank adapters, and \textit{HeteroTune} \cite{jia2024towards} employs prefix-tuning.  These methods cut bandwidth but the adapters remain architecture-specific, so interference resurfaces when scanners, imaging slice thickness, or class priors differ strongly between sites.  Clustering techniques such as \textit{FedGroup} \cite{duan2021fedgroup} mitigate domain shift by grouping clients with similar gradients, but the grouping is computed once at round~0 and cannot react if optimization trajectories later converge or diverge. \textit{FIARSE} introduces importance-aware sub-model extraction and proves an $O(1/\!\sqrt{T})$ convergence rate, yet it still assumes all clients start from a \emph{common} super-network and it updates sub-models by masking weights rather than by a unified parameter space \cite{fiarse}. \textit{FedGLCL} replaces logits with language-image contrastive pairs: a frozen text encoder supplies a global semantic space and each client aligns its image embeddings to that space with CLIP-style loss \cite{yan2025on}.  Although FedGLCL reduces client drift on non-IID data, it presupposes the availability of reliable class prompts and a heavyweight text encoder on the server, and it does not handle architectural conflicts because each client still trains its own image backbone with private parameters.

\texttt{UnifiedFL} removes the need for layer alignment, pruning, or adapter surgery. Each architecture is first converted into a directed model-graph whose nodes and edges are all modulated by a fixed-length GNN parameter vector $\boldsymbol{\Theta}$. Because every client transmits only $\boldsymbol{\Theta}$, tensor shapes never have to match. This stands in sharp contrast to scale-search approaches~\cite{kim2021autofl,diao2020heterofl}, which still rely on a shared backbone and merely resize it—shrinking or deepening layers, channels, or resolutions—to fit each device; such proportional scaling preserves layer-wise averaging but bars truly different architectures. After each communication round, the server computes Euclidean distances between the received $\boldsymbol{\Theta}$ vectors, reclusters clients via Ward linkage, and enforces a two-tier schedule: frequent averaging within clusters and sparse averaging across clusters. Consequently, similar hospitals collaborate often, while highly divergent sites synchronize only after partial convergence—overcoming both the static-cluster constraint of FedGroup \cite{duan2021fedgroup} and the architectural rigidity of scale-search methods.

\paragraph*{Knowledge distillation for heterogeneous models}
An alternative to parameter alignment is to fuse predictions.  \textit{FedMD} \cite{li2019fedmd}, \textit{FedDF} \cite{lin2021ensembledistillationrobustmodel}, and \textit{Cronus} \cite{chang2019cronus} average softened logits on a public data set to train a global student.  Medical imaging rarely offers such a public pool; even when available, privacy regulations may restrict its use.  \textit{MH-pFLID} \cite{xie2024mhpflidmodelheterogeneouspersonalized} eliminates the public set by introducing a “messenger’’ network that visits each client in turn and accumulates knowledge, but the messenger must itself be communicated and trained, adding latency and memory overhead.  Communication-efficient variants compress logits with contrastive objectives \cite{wu2022communication,xing2021categorical}, yet always require two forward passes per batch (teacher and student) and cannot avoid disclosing class-conditional information.

\paragraph*{Unified learning}\label{sec:unified_learning}
\textit{uGNN}~\cite{ugnn} introduces a unified learning paradigm for training heterogeneous neural architectures. Each architecture is transformed into a common graph representation, where weights and biases are encoded as node and edge features of a model graph. A central GNN then performs a custom message-passing procedure over these model graphs to emulate each architecture’s forward pass, enabling knowledge sharing and joint training across models. Rather than optimizing every architecture independently—which is especially challenging when they are trained on data from different distributions—\textit{uGNN} trains only the central GNN. During training, the central GNN learns update rules for node and edge features that mirror the standard SGD-based updates of weights and biases in the original architectures while simultaneously leveraging knowledge shared across architectures, allowing cross-domain information to enhance each model’s individual performance. However, uGNN suffers from a key limitation: it operates in a centralized setting, requiring all model-graphs to be trained in an ensemble learning fashion, which is infeasible in privacy-sensitive domains such as healthcare. Our proposed \texttt{UnifiedFL} framework overcomes this issue by designing a federated setup where clients retain their local data and model-graphs, exchanging only a compact, fixed-length GNN parameter vector $\Theta$ to ensure architecture-agnostic aggregation without sharing full model weights. Furthermore, we introduce a dynamic, $\Theta$-guided clustering mechanism that adaptively groups clients with similar optimization trajectories, thereby mitigating non-IID interference and accommodating architectural drift during training. This combination enables \texttt{UnifiedFL} to deliver the benefits of uGNN’s unified parameter space while ensuring privacy and robustness to both model and data heterogeneity.

% This paper is structured as follows: Section~\ref{sec:preliminary} introduces the necessary preliminaries on federated learning and its heterogeneous settings; Section~\ref{sec:method} presents our proposed method; Section~\ref{sec:experiments} details the experimental setup and results; and Section~\ref{sec:conclusion} concludes the paper with a discussion of findings and future directions. 

\section{Preliminary: Heterogeneous Federated Learning}\label{sec:preliminary}
\textbf{Table}~\ref{tab:notation} summarizes the key mathematical symbols used throughout the paper. We adopt boldface uppercase letters (e.g., \(\mathbf{\Theta}\)) for parameter sets, boldface lowercase letters (e.g., \(\mathbf{v}\)) for vectors, and script fonts (e.g., \(\mathcal{D}_i\)) for datasets or sets of clients.

\label{sec:notation}
\begin{table}[H]
\centering
\caption{Key notation and symbols used in our methodology.}
\label{tab:notation}
\begin{tabular}{ll}
\toprule
\textbf{Symbol} & \textbf{Description} \\
\midrule
\(m\)            & number of clients \\
\(\mathcal{D}_i\)          & local dataset for client \(i\) \\
\(\mathbf{\Theta}\)        & global parameter set (gnn-based) \\
\(\mathbf{\Theta}_{[\mathcal{C}_k]}\) & cluster-level aggregated parameters for cluster \(\mathcal{C}_k\) \\
\(\mathbf{V}_i, \mathbf{E}_i\)        & node and edge feature sets of the model-graph for client \(i\) \\
\(g_e(u,v),\,g_v(v)\)      & group indices for edges and nodes, respectively \\
\(\mathbf{E}_{u,v}\)       & weight from node \(u\) to node \(v\) in a model-graph \\
\(\mathbf{V}_v\)           & bias (node feature) for node \(v\) \\
\(\mathbf{\tau}_i\)        & topology descriptor of client \(i\) \\
\(\mathcal{C}_k\)          & cluster \(k\) of clients \\
\(t_{\mathrm{ic}}, t_{\mathrm{bc}}\)   & communication intervals for intra- and cross-clusters \\
\(t_{\mathrm{init}}\)       & threshold round to begin inter-cluster communication \\
\(t\)                      & total number of federated training rounds \\
\(\mathcal{L}_i\)          & local loss function at client \(i\) \\
\(\sigma(\cdot)\)          & activation function (e.g., relu) \\
\(\mathrm{SoftSign}(\cdot)\) & element-wise softsign operator for scaling and shifting \\
\bottomrule
\end{tabular}
\end{table}

We consider \( K \) clients, where each client \( k \in [K] \) holds a local dataset \(\{(x_i, y_i)\}_{i=1}^{n_k}\), with \((x_i, y_i) \sim \mathbb{P}_k(x, y)\). The goal is to learn a global model \( f \) by solving:
\begin{equation}
\mathcal{L}(\mathbf{w}) = \sum_{k=1}^{K} \gamma_k \mathcal{L}_k(\mathbf{w}), 
\quad \text{where} \quad \gamma_k = \frac{n_k}{\sum_{i=1}^{K} n_i}.
\end{equation}

Heterogeneity in FL arises from two principal sources: \emph{model heterogeneity} and \emph{statistical heterogeneity}~\cite{ye2023heterogeneous}, both of which challenge conventional aggregation schemes such as FedAvg~\cite{mcmahan2017communication}.

\subsection{Model Heterogeneity}
Model heterogeneity arises when clients use different model architectures or parameter spaces due to varying hardware capabilities, software environments, or design choices. In this case, the parameter vector for client \( k \) is denoted \( \mathbf{w}_k \in \mathbb{R}^{p_k} \), where \( p_k \) is specific to client \( k \). Aggregation across heterogeneous models is not straightforward, as parameter vectors may differ in dimension, structure, or semantics, rendering conventional weighted averaging infeasible~\cite{smith2017federated}.

\subsection{Statistical Heterogeneity}
Statistical heterogeneity stems from non-identically distributed (non-IID) data across clients. Each client \( k \) draws data from a client-specific distribution \( \mathbb{P}_k(x, y) \), leading to divergence between local and global optima. The local loss for client \( k \) at round \( t \) is:
\begin{equation}
\mathcal{L}_k(\mathbf{w}_k^t) = \frac{1}{n_k} \sum_{(x_i, y_i) \sim \mathbb{P}_k} \ell\bigl(f(x_i; \mathbf{w}_k^{t-1}), y_i\bigr),
\end{equation}
and the aggregated global update is:
\begin{equation}
\mathbf{w}_G^t = \sum_{k=1}^{K} \gamma_k \mathbf{w}_k^t.
\end{equation}

Mathematically, statistical heterogeneity complicates convergence because local gradients 
\(
\nabla \mathcal{L}_k(\mathbf{w})
\)
tend to point in different directions:
\begin{equation}
\mathbb{E}_{\mathbb{P}_k}\bigl[ \nabla \mathcal{L}_k(\mathbf{w}) \bigr] 
\neq 
\mathbb{E}_{\mathbb{P}}\bigl[ \nabla \mathcal{L}(\mathbf{w}) \bigr],
\end{equation}
where 
\(
\mathbb{P}(x,y) = \sum_{k=1}^K \gamma_k \mathbb{P}_k(x,y)
\)
is the overall population distribution. In other words, local gradient directions are biased toward minimizing their own local losses and may conflict with each other.

The aggregated update can therefore oscillate or fail to make consistent progress:
\begin{equation}
\mathbf{w}_G^{t+1} = \mathbf{w}_G^t - \eta \sum_{k=1}^K \gamma_k \nabla \mathcal{L}_k(\mathbf{w}_G^t),
\end{equation}
where the sum of gradients may not approximate the true global gradient 
\(
\nabla \mathcal{L}(\mathbf{w}_G^t)
\)
well. This misalignment slows down convergence and may even lead to divergence if client distributions are highly dissimilar~\cite{li2020federated,li2020scaffold,li2022federated}.

% \subsection{\textbf{Graph-Based Unified Parameter Space}}
% \label{sec:graph_parameter_space}

% \textbf{Problem definition.}
% Hospitals train different, evolving backbones on non-IID images; direct averaging fails because weight shapes disagree and their gradient directions conflict.  
% How can we federate these models without sharing raw data by \emph{(i)} replacing weight-wise aggregation with a common graph parameter $\boldsymbol{\Theta}$ and \emph{(ii)} dynamically clustering clients by their current $\boldsymbol{\Theta}$ to decide when and with whom to synchronize?

\section{Proposed \texttt{UnifiedFL}}\label{sec:method}

\textbf{Problem formulation.}  
Consider an FL system with $m$ hospitals indexed by $k\in[K]$.  
Each hospital stores a private image set $\mathcal{D}_{k}$ sampled from an unknown distribution $\mathbb{P}_{k}(x,y)$ that may differ across sites (\emph{non-IID}).  
Hospital~$k$ chooses an architecture $f_{k}(\cdot\,;\mathbf{W}_{k})$—e.g.\ CNN, U-Net, or MLP—whose trainable weights and biases are collected in $\mathbf{W}_{k}\in\mathbb R^{d_k}$ with \emph{architecture-specific} dimensionality~$d_k$.  
Directly averaging the $\mathbf{W}_{k}$ is impossible because each weight vector has different dimensions.

\begin{figure}[!h]
    \centering
    \includegraphics[width=\linewidth]{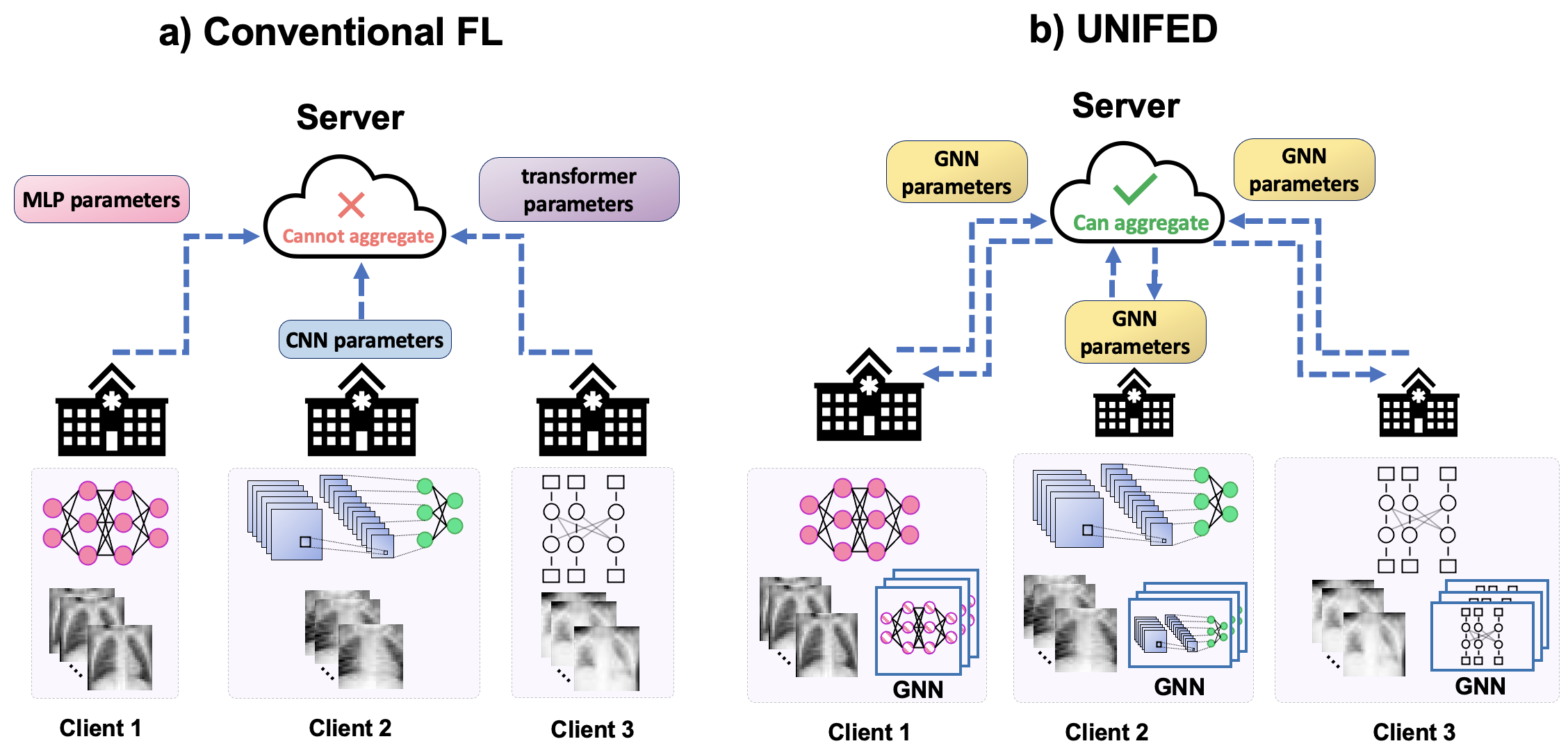}
\caption{Conceptual comparison between conventional federated learning and the proposed unified learning (\texttt{UnifiedFL}) workflow.  
\textbf{(a) Conventional FL.}  
Clients deploy heterogeneous backbones (MLP, CNN, Transformer).  
Each site uploads its native weight tensor to the server (blue dashed arrows).  
Because the tensors differ in shape, the server cannot perform element-wise aggregation (red cross).  
\textbf{(b) \texttt{UnifiedFL}.}  
Each client converts its backbone to a model-graph and trains a shared set of GNN parameters~$\boldsymbol{\Theta}$ that rescale the underlying weights.  
Only the compact $\boldsymbol{\Theta}$ is exchanged.  
All parameter vectors have identical length, so the server can average them directly (green tick) and broadcast the result back to the hospitals.  
This mechanism enables architecture-agnostic collaboration without exposing raw images or full model weights.}

    \label{fig:concept_figure}
\end{figure}

To enable aggregation we convert client architecture backbone into a \emph{model-graph} $\mathcal{G}_{k}=(\mathbf{V}_{k},\mathbf{E}_{k})$. Let $\boldsymbol{\Theta}\!\in\!\mathbb{R}^{p}$ denote the \emph{shared} GNN parameter vector that updates every edge feature $\mathbf{E}_{u,v}$ and node bias $\mathbf{V}_{v}$ in every client-side model-graph.  
Optimizing $\boldsymbol{\Theta}$ therefore \emph{indirectly} tunes the underlying weights of \textit{all} heterogeneous backbones.  
We wish to discover a single optimum  
$\tilde{\boldsymbol{\Theta}}$ that minimizes the mean empirical loss across hospitals,  
\[
\tilde{\boldsymbol{\Theta}}
=
\argmin_{\boldsymbol{\Theta}}
F(\boldsymbol{\Theta}),
\quad
F(\boldsymbol{\Theta})
\;=\;
\frac{1}{K}\sum_{i=1}^{K} F_{i}(\boldsymbol{\Theta}),
\]
where $F_{i}(\boldsymbol{\Theta})$ is the loss on the private set $\mathcal{D}_{i}$ after updating $(\mathbf{V}_{i},\mathbf{E}_{i})$ with $\boldsymbol{\Theta}$.

% and controll all node and edge features with a \emph{shared} GNN parameter vector $\boldsymbol{\Theta}\in\mathbb{R}^{p}$ whose length $p$ is identical for all clients.  
% Let $g_{i}(\mathbf{V}_{i},\mathbf{E}_{i};\boldsymbol{\Theta})$ denote the forward pass through $\mathcal{G}_{i}$.
% The local empirical loss at hospital $i$ is
% \[
% F_{i}(\boldsymbol{\Theta})\;=\;
% \frac{1}{|\mathcal{D}_{i}|}\,
% \sum_{(x,y)\in\mathcal{D}_{i}}
% \ell\Bigl(
% g_{i}(\mathbf{V}_{i},\mathbf{E}_{i};\boldsymbol{\Theta})(x),\,y
% \Bigr),
% \]
% where $\ell$ is the task loss, e.g., cross-entropy for classification.

\paragraph{Dynamic $\boldsymbol{\Theta}$-guided clustering}  
After each local epoch hospital $i$ holds an updated copy  
$\boldsymbol{\Theta}^{[t]}_{(i)}$.  
The server forms a distance matrix  
$D_{ij}^{[t]}=\Vert\boldsymbol{\Theta}^{[t]}_{(i)}-\boldsymbol{\Theta}^{[t]}_{(j)}\Vert_{2}$,  
applies Ward’s linkage hierarchical clustering, and obtains a  partition  
$\mathcal{C}^{[t]}=\{\mathcal{C}_{1}^{[t]},\dots,\mathcal{C}_{M}^{[t]}\}$ with $M$ clusters.
Hence clients that follow similar optimization trajectories are grouped together, while diverging ones are separated.

\paragraph{Two–level aggregation schedule}  
Given $\mathcal{C}^{[t]}$ we perform

\[
\boldsymbol{\Theta}^{[t]}_{[\mathcal{C}_{m}]}
=
\frac{1}{|\mathcal{C}_{m}^{[t]}|}\!
\sum_{i\in\mathcal{C}_{m}^{[t]}}
\boldsymbol{\Theta}^{[t]}_{(i)}
\quad\text{(every }t_{\mathrm{ic}}\text{ rounds)},
\]

\[
\boldsymbol{\Theta}^{[t]}
=
\frac{1}{M}
\sum_{m=1}^{M}
\boldsymbol{\Theta}^{[t]}_{[\mathcal{C}_{m}]}
\quad\text{(every }t_{\mathrm{bc}}>t_{\mathrm{ic}}\text{ rounds)},
\]

where $\boldsymbol{\Theta}^{[t]}_{[\mathcal{C}_{k}]}$ is the intra-cluster average and $\boldsymbol{\Theta}^{[t]}$ is the global vector broadcast for the next round.  
Frequent intra-cluster exchange accelerates convergence among similar hospitals; infrequent inter-cluster exchange prevents destructive interference until models have partially aligned.  
Because only the \emph{fixed-length} vectors $\boldsymbol{\Theta}^{[t]}_{(i)}$ are transmitted—never raw images or backbone-specific weights—the procedure respects privacy while jointly optimizing heterogeneous, dynamically evolving architectures.

% \textbf{Problem definition.}
% Hospitals train heterogeneous models on non-IID image data, making conventional aggregation ineffective due to mismatched weight shapes and conflicting gradient directions.
% How can we enable federated learning without sharing raw data by:
% \emph{(i)} replacing direct weight aggregation with a shared graph-based parameter $\boldsymbol{\Theta}$, and
% \emph{(ii)} dynamically clustering clients based on their current $\boldsymbol{\Theta}$ to determine when and with whom to synchronize?

% \subsection{\textbf{Notation and symbol table}}

\begin{figure}[t]
    \centering
    \includegraphics[width=\linewidth]{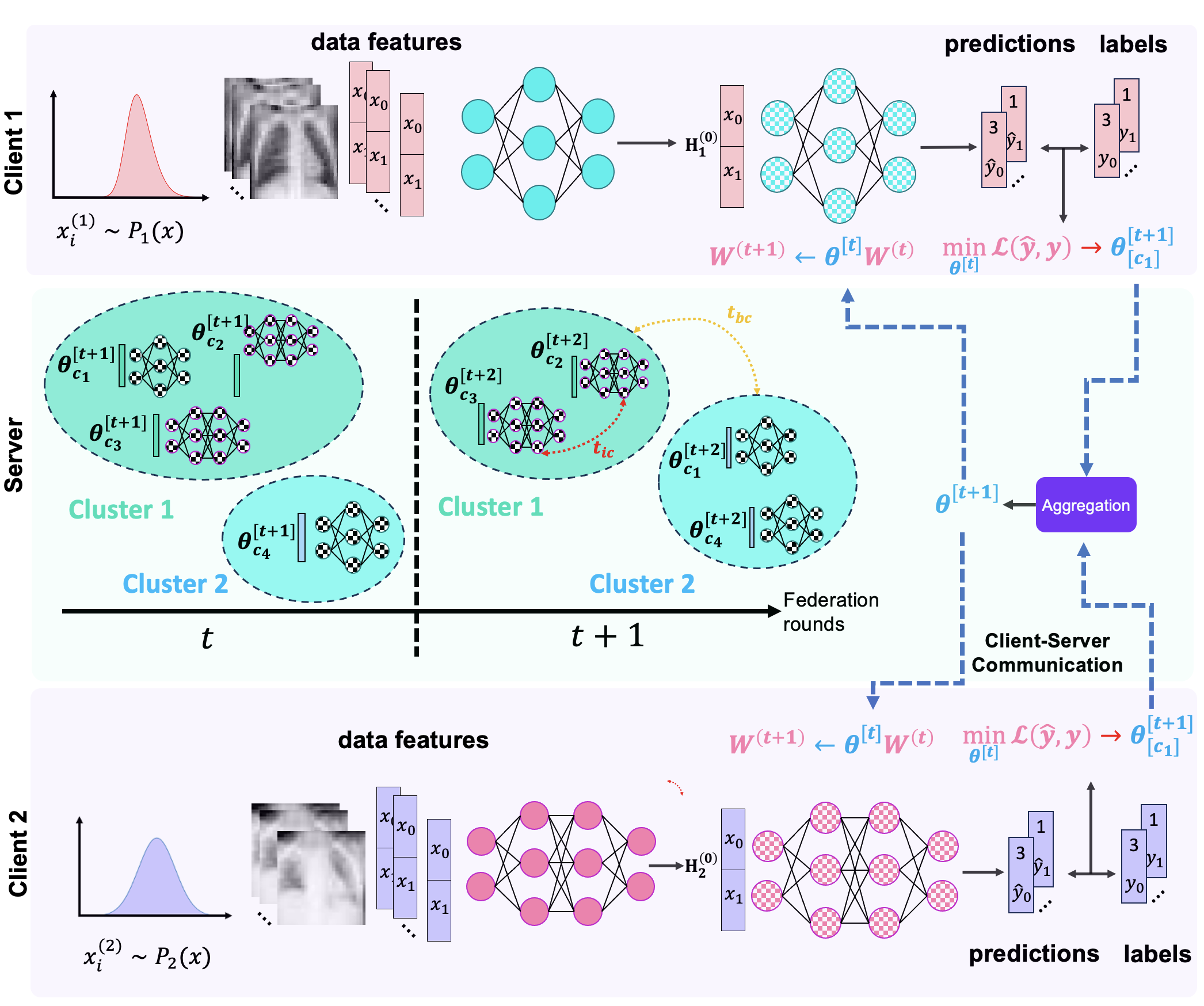}
\caption{%
\textbf{Overview of the proposed \texttt{UnifiedFL} workflow.}
At every federation round~$t$ each client (top \& bottom rows show examples for two hospitals with distinct image distributions $P_{1}(x)$ and $P_{2}(x)$) converts its private backbone into a \emph{model-graph}, optimizes the shared GNN parameters~$\boldsymbol\theta^{[t]}$ on local data and sends the updated parameters $\boldsymbol\theta^{[t+1]}_{[c_i]}$ (blue dashed arrows) rather than raw network weights~$\mathbf W$.
The server (center) clusters clients according to graph topology; frequent \emph{intra-cluster} aggregations every $t_{\mathrm{ic}}$ (dotted red arrows) are complemented by sparser \emph{inter-cluster} merges every $t_{\mathrm{bc}}$ (dotted yellow arrows).
This topology-aware schedule prevents interference between dissimilar architectures while still enabling global knowledge transfer.
Aggregated parameters are broadcast back to all clients, where they rescale/shift local weights for the next round, yielding an architecture-agnostic and communication-efficient federated learning process.}

    \label{fig:main_fig}
\end{figure}

\textbf{Our motivation.}
\textbf{Fig.}~\ref{fig:concept_figure}a shows a typical failure mode of FedAvg when hospitals deploy dissimilar backbones.  
Client 1 trains an MLP, Client 2 trains a CNN, and Client 3 trains a Transformer.  
The weight tensors uploaded to the server differ in rank and spatial layout, so the server cannot perform element-wise aggregation.  
\textbf{Fig.}~\ref{fig:concept_figure}b outlines the \texttt{UnifiedFL} remedy.  
Each backbone is rewritten as a directed acyclic \emph{model-graph}: nodes hold biases or spatial activations, and edges hold convolutional or linear weights.  
All clients then optimize a shared GNN parameter vector~$\boldsymbol{\Theta}$ that rescales these node and edge features.  
Because every $\boldsymbol{\Theta}$ has identical length, the server can average updates directly and broadcast the result without shape conflicts.  
The complete data-flow, including intra- and inter-cluster aggregation, is detailed in \textbf{Fig.}~\ref{fig:main_fig}. Building on this motivation, we formulate the following hypotheses that underpin the design of our proposed framework:
\\
\textbf{H1.} A backbone-agnostic representation in which every local network is rewritten as a model-graph and updated solely through a shared parameter vector $\boldsymbol{\Theta}$ is sufficient to remove all tensor-shape barriers to aggregation.\\[2pt]
\textbf{H2.} Measuring pair-wise Euclidean distances between the current $\boldsymbol{\Theta}$ vectors and reclustering clients at each round yields communication groups that track optimization similarity and therefore reduce the gradient conflict induced by non-IID data.\\[2pt]
\textbf{H3.} Combining H1 and H2 allows federated training to approach the accuracy of a centralized oracle while keeping raw images and full model weights strictly on site.

In the following subsections, we detail the building blocks of \texttt{UnifiedFL}.

\subsection{Model-graph consturction and unification}
Given a client \(i\) with a local network of arbitrary design, we define
$
\mathbf{V}_i = \{ \mathbf{V}_v : v \in \mathcal{N}_i \}, 
\quad
\mathbf{E}_i = \{ \mathbf{E}_{u,v} : (u \to v) \in \mathcal{A}_i \},
$
where \(\mathcal{N}_i\) is the set of nodes (neurons or feature-map positions) and \(\mathcal{A}_i\) is the set of directed edges (weights). This transformation does not require layer-by-layer alignment; hence the architecture is preserved in a flexible graph form \cite{kofinas2024graph}.

\textbf{Unified GNN parameters.}
Rather than exchanging full local models, our framework introduces a global GNN-based parameter set \(\mathbf{\Theta}\) that \emph{indirectly} updates local weights (\(\mathbf{E}_{u,v}\)) and biases (\(\mathbf{V}_v\)). As shown in \textbf{Fig.}~\ref{fig:main_fig}\,(b), \(\mathbf{\Theta}\) is partitioned into \(\{\mathbf{\Theta}_{\mathrm{node}}, \mathbf{\Theta}_{\mathrm{edge}}\}\) with additional shift and scale parameters. Specifically, edges and nodes are grouped, and each group is associated with a scale/shift pair \((\mathbf{\Theta}_{\mathrm{edge}}, \mathbf{\Theta}_{\mathrm{edge\_shift}})\) or \((\mathbf{\Theta}_{\mathrm{node}}, \mathbf{\Theta}_{\mathrm{node\_shift}})\). Let \(g_e(u,v)\) be the group assignment for edge \((u\to v)\) and \(g_v(v)\) the group for node \(v\). Next, we update the edge and node embeddings as follows:
\[
\mathbf{E}_{u,v} \leftarrow 
\mathrm{SoftSign}\Bigl(
\mathbf{E}_{u,v}\,\mathbf{\Theta}^{(g_e(u,v))}_{\mathrm{edge}}
+ \mathbf{\Theta}^{(g_e(u,v))}_{\mathrm{edge\_shift}},
\,\mathbf{\Theta}_{\mathrm{scale\_edge}}
\Bigr),
\]
\[
\mathbf{V}_{v} \leftarrow 
\mathrm{SoftSign}\Bigl(
\mathbf{V}_{v}\,\mathbf{\Theta}^{(g_v(v))}_{\mathrm{node}}
+ \mathbf{\Theta}^{(g_v(v))}_{\mathrm{node\_shift}},
\,\mathbf{\Theta}_{\mathrm{scale\_node}}
\Bigr).
\]

\subsection{Federated optimization and static clustering}
\label{sec:static_clustering}

\textbf{Local feedforward and loss.}
Once each node and edge feature is updated by \(\mathbf{\Theta}\), the feedforward pass of the local model-graph proceeds as:
\[
\mathbf{H}_{v}^{(\ell)} \;=\; \sigma\!\Bigl[\!\!\sum_{u : \mathbf{A}_{u,v}=1}
\bigl(\mathbf{E}_{u,v}\,\mathbf{H}_{u}^{(\ell-1)} \;+\; \mathbf{V}_{v}\bigr)\Bigr],
\]
where \(\mathbf{H}_{v}^{(\ell)}\) denotes the activations at node \(v\) in layer \(\ell\), and \(\sigma(\cdot)\) is an activation function. A local loss \(\mathcal{L}_i(\mathbf{\Theta})\) compares the outputs of the model-graph with local labels in dataset \(\mathcal{D}_i\).

\textbf{Federated averaging.}
Under a standard FL setting, the server broadcasts \(\mathbf{\Theta}^{[t]}\) to all clients each round \(t\). Client \(i\) performs local gradient steps:
\[
\mathbf{\Theta}_{[i]}^{[t]} \;=\; \mathbf{\Theta}^{[t]} \;-\; \eta \,\nabla_{\mathbf{\Theta}}\,\mathcal{L}_i(\mathbf{\Theta}^{[t]}) \quad\forall i,
\]
and the server aggregates via weighted averaging:
\[
\mathbf{\Theta}^{[t+1]} \;=\; \frac{1}{m}\,\sum_{i=1}^m \mathbf{\Theta}_{[i]}^{[t]}.
\]

\textbf{Topology-aware static clustering.}
Each client computes a \emph{topology descriptor} \(\mathbf{\tau}_i\) (e.g., node degrees, betweenness centrality) once at initialization. The server forms clusters \(\{\mathcal{C}_1, \dots, \mathcal{C}_K\}\) based on these descriptors and maintains them \emph{throughout} training. Clients in the same cluster communicate more frequently (\(t_{\mathrm{ic}}\)) than clients in different clusters (\(t_{\mathrm{bc}}\)). Although effective in reducing interference, this static clustering is oblivious to \emph{dynamic changes} in learned parameters and architectural adaptations. We treat this static-clustering design as an \emph{ablation} of our new method (\texttt{UnifiedFL}), since it relies purely on initial topological features.

\subsection{Dynamic $\boldsymbol{\Theta}$-guided clustering}
\label{sec:dynamic_clustering}

\textbf{Distance metric.}
At federation round $t$ each client $i$ holds an updated copy of the shared GNN parameters $\boldsymbol{\Theta}^{[t]}_{(i)}\in\mathbb{R}^{P}$.  
We construct a symmetric distance matrix
\[
D_{ij}^{[t]}
    \;=\;
    \lVert
        \boldsymbol{\Theta}^{[t]}_{(i)}
      - \boldsymbol{\Theta}^{[t]}_{(j)}
    \rVert_{2},
\quad
1\!\le i,j\!\le m ,
\]
which measures the instantaneous $\ell_2$ divergence of optimization states across sites. This procedure relies solely on model parameters, making it a \emph{parameter-only} approach that directly captures real-time learning trajectories without auxiliary statistics such as gradient dispersion or graph topological descriptors.

\begin{algorithm}[H]

\caption{ \texttt{UnifiedFL} (Proposed) with \textbf{Dynamic Clustering}}
\label{alg:UnifiedFL_algorithm}
\begin{algorithmic}[1]
\Require Number of clients $m$, local data $\{\mathcal{D}_i\}$, total rounds $T$, intervals $t_{\mathrm{ic}}, t_{\mathrm{bc}}, t_{\mathrm{update}}$, initial $\mathbf{\Theta}^{[0]}$.
\Ensure Final global model $\mathbf{\Theta}^{[T]}$, local model-graphs $(\mathbf{V}_i,\mathbf{E}_i)$
\State \textbf{Initialization:} Each client $i$ constructs $(\mathbf{V}_i,\mathbf{E}_i)$ and calculates initial descriptor $\mathbf{\tau}_i^{[0]}$ (topology + partial param).
\State Server clusters the clients into $\{\mathcal{C}_1, \dots, \mathcal{C}_K\}$ based on $\mathbf{\tau}_i^{[0]}$.
\For{$t = 1$ to $T$}
    \State Server broadcasts $\mathbf{\Theta}^{[t-1]}$.
    \For{each client $i$}
        \State $\mathbf{\Theta}_{[i]}^{[t]} \gets \mathbf{\Theta}^{[t-1]} - \eta \,\nabla_{\mathbf{\Theta}} \mathcal{L}_i\bigl(\mathbf{\Theta}^{[t-1]}\bigr)$
    \EndFor
    \If{$t \bmod t_{\mathrm{ic}} = 0$}
        \For{each cluster $\mathcal{C}_k$}
            \State $\mathbf{\Theta}_{[\mathcal{C}_k]}^{[t]} \gets \frac{1}{|\mathcal{C}_k|}\,\sum_{i \in \mathcal{C}_k} \mathbf{\Theta}_{[i]}^{[t]}$
        \EndFor
    \EndIf
    \If{$t > T_{\mathrm{init}}$ and $t \bmod t_{\mathrm{bc}} = 0$}
        \State $\mathbf{\Theta}^{[t]} \gets \frac{1}{K} \sum_{k=1}^K \mathbf{\Theta}_{[\mathcal{C}_k]}^{[t]}$
    \EndIf
    \If{$t \bmod t_{\mathrm{update}} = 0$} 
        \Comment{Update clustering dynamically}
        \For{each client $i$}
            \State \(\mathbf{\tau}_i^{[t]} \gets \bigl[\mathbf{\tau}_i^\mathrm{(topo)},\, \mathbf{\tau}_i^\mathrm{(param\,@\,t)}\bigr]\)
            \Comment{Combine topology \& current parameters}
        \EndFor
        \State Server re-clusters $\{\mathcal{C}_1, \dots, \mathcal{C}_K\}$ using $\{\mathbf{\tau}_i^{[t]}\}_{i=1}^m$
    \EndIf
\EndFor
\State \Return $\mathbf{\Theta}^{[T]}$, $(\mathbf{V}_i,\mathbf{E}_i)$
\end{algorithmic}
\end{algorithm}

\textbf{Hierarchical clustering.}
Using $D^{[t]}$ we perform agglomerative clustering with Ward’s linkage.  
The linkage tree is cut at the level that maximizes the average silhouette score; hence the number of clusters $K^{[t]}$ is data-driven and may vary with~$t$.  
All cluster assignments are recomputed \emph{every} round, so $\mathcal{C}_k^{[t]}$ can evolve without inertia.

\textbf{Communication schedule.} After clustering, the server applies two aggregation rates: \textbf{(1) Within-cluster} synchronization every $t_{\mathrm{ic}}$ rounds.  
\textbf{(2) Between-cluster} synchronization every $t_{\mathrm{bc}}$ rounds, with $t_{\mathrm{bc}}>t_{\mathrm{ic}}$.  
      During the warm-up period $t<T_{\mathrm{init}}$ we set $t_{\mathrm{bc}}=\infty$, i.e.\ no cross-cluster exchange. This staggered schedule lets similar models share updates frequently while shielding dissimilar models early in training.  Later, sporadic cross-cluster exchange promotes global consensus without imposing strong interference.

\textbf{Complexity and privacy.}
Forming $D^{[t]}$ costs $\mathcal{O}(m^2P)$ additions, negligible for $m\le10$ and $P\approx4\times10^{5}$.  
Only Euclidean distances are revealed to the server; the raw $\boldsymbol{\Theta}^{[t]}_{(i)}$ remain local, preserving parameter privacy.
Algorithm~\ref{alg:UnifiedFL_algorithm} lists one training session of \texttt{UnifiedFL}.  
After a one-off graph conversion, each hospital holds a private model-graph $(\mathbf{V}_i,\mathbf{E}_i)$ and a copy of the shared GNN parameters $\mathbf{\Theta}^{[0]}$.  
At the beginning of every federation round the server broadcasts the current $\mathbf{\Theta}^{[t-1]}$.  
Each client performs one local epoch of stochastic gradient descent on its own data, producing an updated parameter vector $\mathbf{\Theta}^{[t]}_{(i)}$.  

\textbf{Stage~1---within-cluster merge.}  
Every $t_{\mathrm{ic}}$ rounds the server averages $\mathbf{\Theta}^{[t]}_{(i)}$ inside each cluster $\mathcal{C}_k$ to obtain a cluster centers $\mathbf{\Theta}^{[t]}_{[\mathcal{C}_k]}$.  
This frequent synchronization transfers knowledge only among models that the current clustering deems similar, thereby reducing destructive interference.

\textbf{Stage~2---between-cluster merge.}  
After an initial warm-up of $T_{\mathrm{init}}$ rounds, the server performs a slower cross-cluster merge every $t_{\mathrm{bc}}$ rounds, averaging the cluster centres to refresh the global parameters $\mathbf{\Theta}^{[t]}$.  
All clients then replace their local copy with this global vector.

\textbf{Stage 3---dynamic re-clustering.}  
Every $t_{\mathrm{update}}$ rounds each client transmits a compact eight-dimensional descriptor of the first and second moments of its per-group gradients with respect to $\mathbf{\Theta}$.  
Using these descriptors the server recomputes the pair-wise Euclidean distance matrix, applies Ward’s hierarchical clustering, and updates the partition $\{\mathcal{C}_k\}$.  
Since only gradient statistics are shared, no model weights or images leave the local clients.

The three stages repeat until the prescribed number of federation rounds $T$ is reached.  
The algorithm terminates with a single global parameter vector $\mathbf{\Theta}^{[T]}$ and a tuned model-graph at every hospital.  
Communication cost per round is $\mathcal{O}(|\mathbf{\Theta}|)$ floats, and the extra cost of the descriptors is fixed at 32 bytes per client every $t_{\mathrm{update}}$ rounds.

% \subsection{Ablation: \texttt{uGNN} and \texttt{uFedGNN} as  baselines}
% \label{sec:ablation_ufedgnn}

% Our proposed \textbf{UnifiedFL} subsumes \textbf{uFedGNN} as an ablated version in which dynamic clustering is \emph{disabled}. Specifically, uFedGNN uses the same graph-based parameter space but relies solely on an \emph{initial, static clustering} derived from topological descriptors. Once groups are formed, no re-clustering occurs throughout training. As shown in our experiments, this ablation can be effective but is less adaptable when client models evolve or their learned parameters diverge. By contrast, UnifiedFL continuously updates cluster memberships, thereby improving robustness in real-world, shifting environments.

\section{Experiments and Results}
\label{sec:experiments}

This section details the experimental protocol used to assess the proposed \texttt{UnifiedFL} framework.  We describe the datasets and pre‑processing pipelines, the strategy used to partition data among clients, the heterogeneous model zoo deployed at each local client, the hyper‑parameters governing training and communication, and the benchmarking measures and computational cost.  

\subsection{Evaluation datasets}
\label{sec:datasets}
We evaluate \texttt{UnifiedFL} on three classification datasets from the \emph{MedMNIST} collection~\cite{medmnistv1,medmnistv2}, one morphology-augmented variant of MNIST, and one 3-D segmentation dataset from the \emph{Medical Segmentation Decathlon} (MSD)~\cite{simpson2019largeannotatedmedicalimage}. \textbf{MorphoMNIST} (70{,}000 grayscale digits, 10 classes) extends the original MNIST by applying elastic deformations that amplify morphological variability, enabling evaluation of a model’s sensitivity to subtle structural differences. From \emph{MedMNIST}, we select \textbf{PathMNIST} (107{,}180 \(32\times32\) RGB tiles, 9 classes), comprising haematoxylin-and-eosin–stained colorectal cancer tissue patches, which form a fine-grained histopathology classification task; \textbf{BreastMNIST} (780 \(28\times28\) grayscale ultrasound images, binary labels), focusing on benign vs. malignant breast lesion detection; and \textbf{PneumoniaMNIST} (5{,}856 chest X-ray crops, binary labels), aimed at paediatric pneumonia diagnosis.  For voxel-level prediction, we include the \textbf{Hippocampus} dataset~\cite{antonelli2022medical} from MSD, consisting of 263 T1-weighted MRI volumes with manual annotations of anterior and posterior hippocampal sub-regions. All volumes are resampled to \(1\,\mathrm{mm}^3\) isotropic resolution, cropped to the hippocampal bounding box, and intensity-normalized to zero mean and unit variance. For all datasets, we preserve the official training–validation–test splits and report metrics exclusively on the held-out test sets. Images are normalized to the range \([0,1]\), with no additional data augmentation applied.

\begin{table}[H]
\centering

\caption{Evaluation datasets.  “Res.” denotes original in‐plane resolution; “G” grayscale; “RGB” three‐channel; H\&E: haematoxylin–eosin stain; “US” ultrasound; “CXR” chest X-ray.  
Counts are number of subjects or images ($\text{k}=10^{3}$).  
“cls.” stands for classification; “seg.” for segmentation.}
\label{tab:data_stats}
\begin{tabular}{lccccc}
\toprule
\rowcolor{gray!10}
\textbf{Dataset} & \textbf{Modality / res.} & \textbf{Task} & \textbf{Classes} & \textbf{Train/Val/Test} \\
\midrule
MorphoMNIST      & Synth.~digits, $28^2$ G  & 2-D cls. & 10  & 60k / 10k / 10k \\
PathMNIST        & H\&E, $32^2$ RGB         & 2-D cls. & 9   & 90k / 10k / 7.2k \\
BreastMNIST      & US, $28^2$ G             & 2-D cls. & 2   & 546 / 78 / 156  \\
PneumoniaMNIST   & CXR, $28^2$ G            & 2-D cls. & 2   & 4.7k / 0.5k / 0.6k \\
Hippocampus (MSD)& T1 MRI, $1\text{mm}^3$   & 3-D seg. & 2  & 211 / 32 / 20   \\
\bottomrule
\end{tabular}
\end{table}

\subsection{Data clustering}
\label{sec:clustering}

\begin{figure}[H]
    \centering
    \includegraphics[width=\linewidth]{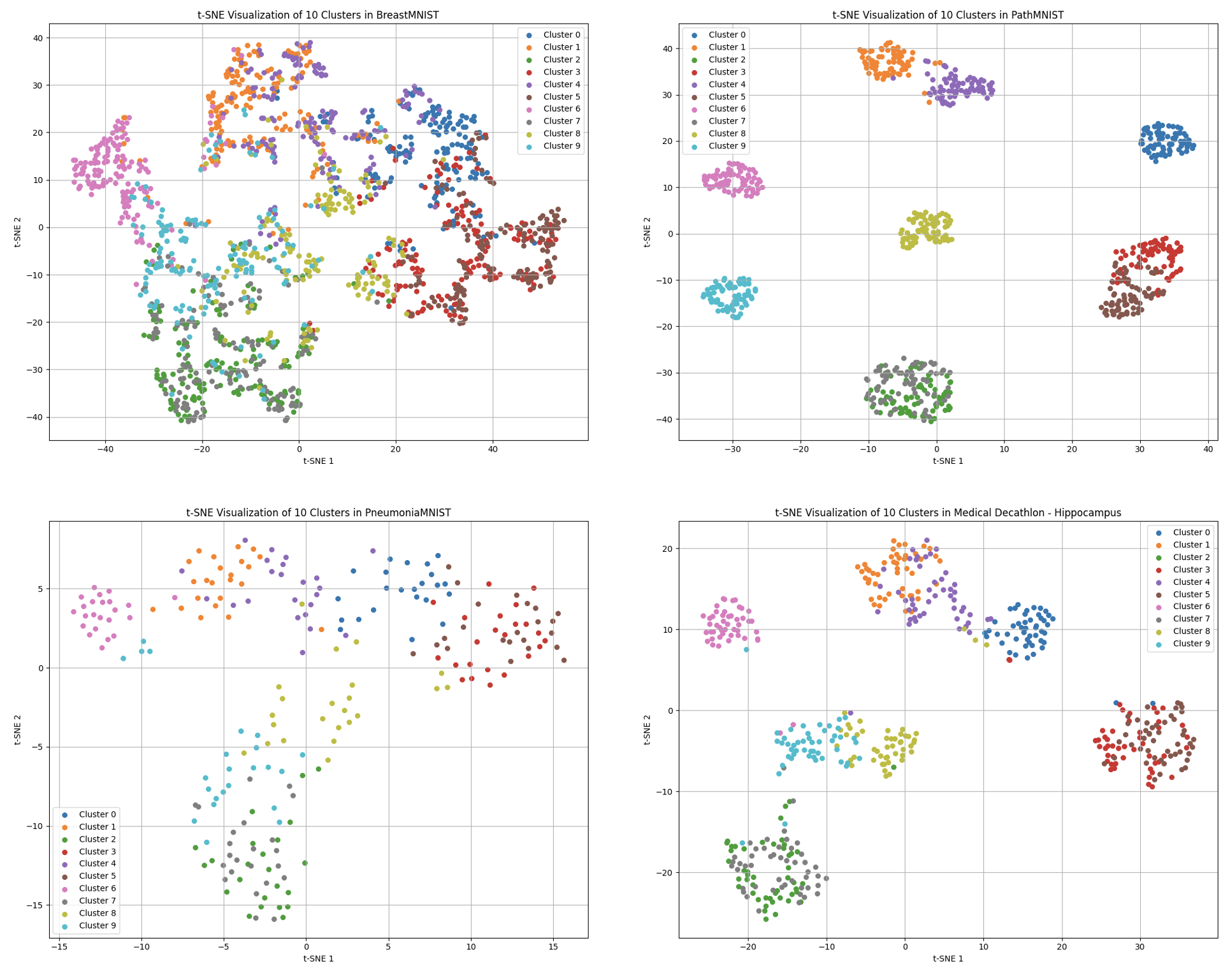}
    \caption{t-SNE visualisations of the raw feature space used to create non-IID client splits.  
    We run $k$-means with $k{=}10$, and project the features to two dimensions for display.  
    Points sharing color belong to the same $k$-means cluster and will be assigned to the same federated client.  
    The four panels correspond to BreastMNIST, PathMNIST, PneumoniaMNIST, and Hippocampus (clock-wise from top left).  
    Well-separated color clouds indicate strong inter-cluster heterogeneity, whereas overlap signals milder shifts; these visual patterns anticipate the non-IID difficulty faced during federated training.}
    \label{fig:tsne_clusters}
    \label{fig:tsne}
\end{figure}

To emulate the severe distribution shifts that arise when hospitals operate different scanners or serve distinct patient populations, we adopt and extend the feature‑based clustering protocol of \texttt{uFedGNN}. Specifically, we first extract 128‑dimensional embeddings with a ResNet‑18 pretrained on ImageNet.  We then apply \(k\)-means (\(k=m\)) to these embeddings and assign every cluster to a separate federated client, yielding strongly non‑IID splits in which diagnostic prevalence, acquisition modality, and image style vary markedly across sites.  For the Hippocampus volumes we compute global intensity histograms concatenated with 18‑dimensional shape descriptors derived from signed distance transforms, and cluster these vectors with Ward’s linkage.  To provide a milder baseline we also form IID splits via uniform random sampling, keeping the per‑client sample size equal to the non‑IID case; this helps disentangle the effects of architectural heterogeneity from statistical heterogeneity.

\subsection{Ablation study \& benchmark methods}

\paragraph*{Experimental setups}
We evaluate \texttt{UnifiedFL} under two regimes: \emph{(i) fully heterogeneous architectures} and \emph{(ii) partially heterogeneous architectures}, reporting the \emph{local} (per-client) performance in both cases. \textbf{(i) Fully heterogeneous.} Each client is randomly assigned one of ten architectures drawn from \textbf{Table}~\ref{tab:model_stats}. We compare four training strategies: (a) \texttt{UnifiedFL}, (b) its ablated variant \texttt{uFedGNN}, (c) \textit{uGNN}, and (d) single-site training (clients trained independently). Per-client results are summarized in \textbf{Fig.}~\ref{fig:results_overview} and detailed in the Appendix (\textbf{Tables}~\ref{tab:breastmnist}--\ref{tab:hippocampus}). \textbf{(ii) Partially heterogeneous.} To compare with closely related heterogeneous FL baselines that require layer-wise compatibility, we instantiate a moderately heterogeneous cohort using VGG~\cite{simonyan2015deepconvolutionalnetworkslargescale} variants with different depths and parameter counts (VGG11, VGG13, VGG16-C, VGG16-D, VGG19). We benchmark \texttt{UnifiedFL} and \texttt{uFedGNN} against the four heterogeneous FL state-of-the-art baselines listed in \textbf{Table}~\ref{tab:qual_compare}, reporting the average per-client performance in \textbf{Table}~\ref{tab:hetero_perf}. This restricted setup is necessary because the compared methods do not support fully heterogeneous federations (e.g., CNNs mixed with MLPs).

\begin{table}[!h]
\centering

\caption{Architectural complexity of the heterogeneous backbones.  
Layer count refers to trainable layers. For CNN and U-Net, layer count represents number of convolution layers + number of FC layers. Parameter totals are rounded to the nearest $10^{3}$; “M” denotes millions.}
\label{tab:model_stats}
\begin{tabular}{lcc}
\toprule
\rowcolor{gray!10}
\textbf{Model} & \textbf{\# layers} & \textbf{Params} \\
\midrule
CNN\textsubscript{a} & 4+1 & $0.63$M \\
CNN\textsubscript{b} & 8+1 & $3.15$M \\
CNN\textsubscript{c} & 12+1 & $9.70$M \\
U‐Net & 23+1 & $11.3$M \\
MLP\textsubscript{a} & 2 & $0.054$M \\
MLP\textsubscript{b} & 3 & $0.14$M \\
MLP\textsubscript{c} & 4 & $0.30$M \\
MLP\textsubscript{d} & 6 & $0.80$M \\
MLP\textsubscript{e} & 8 & $2.10$M \\
MLP\textsubscript{f} & 10 & $4.40$M \\
\bottomrule
\end{tabular}
\end{table}

\paragraph*{\texttt{uFedGNN} as an ablated version of \texttt{UnifiedFL}} 
To contextualize the contributions of our proposed \texttt{UnifiedFL} framework, we compare it against an ablated version called \texttt{uFedGNN} which serves as a representative baseline. Similar to \textit{uGNN}~\cite{ugnn}, \texttt{uFedGNN} tackles architectural and statistical heterogeneity by projecting diverse neural networks into a shared graph-based parameter space. Specifically, they convert each model into a directed acyclic model-graph, consistent with the approach used in \texttt{UnifiedFL}. As mentioned in Sec.~\ref{sec:unified_learning},  \textit{uGNN} does not operate in a federated setting. Instead, all model-graphs are centrally collected and embedded in a global graph space, where a shared GNN performs forward passes and directly updates node and edge embeddings. These embeddings correspond to the biases and weights of the original neural networks. As all updates are performed centrally by the global GNN, there is no local training or data privacy consideration in \textit{uGNN}. In contrast, \texttt{uFedGNN} adopts a federated learning paradigm in which clients retain their private data and model-graphs locally. A single set of global GNN parameters, denoted by $\mathbf{\Theta} = \{\mathbf{\Theta}_{\text{edge}}, \mathbf{\Theta}_{\text{node}}\}$, governs the update process across all clients. Each client locally optimizes $\mathbf{\Theta}$ on its own data and transmits the updated parameters to the server, which aggregates them to form the new global model. This architecture-agnostic framework avoids explicit layer-wise alignment and significantly reduces communication overhead by sharing only the compact set of GNN parameters, rather than full model weights or architectures.

To mitigate parameter interference between fundamentally different architectures, \texttt{uFedGNN} introduces a topology-aware clustering mechanism. At initialization, each client computes a topological descriptor of its model-graph—such as node degrees or centrality measures—and sends it to the server. Based on these descriptors, the server partitions clients into clusters that remain fixed throughout training. Clients within the same cluster synchronize their GNN parameters more frequently (every $t_{\mathrm{ic}}$ rounds), while cross-cluster synchronization is deferred until a later stage of training (after round $T_{\mathrm{init}}$), and occurs less frequently (every $t_{\mathrm{bc}}$ rounds). This strategy aims to allow similar models to collaborate more often, while limiting interference from dissimilar ones.

Despite its innovation, \texttt{uFedGNN} has an inherent limitation: its clustering mechanism is static and topology dependent. In practical federated environments, local models may evolve during training—adapting layers, changing hyperparameters, or learning parameters that diverge or converge significantly across time. Fixed clusters cannot capture such dynamic shifts in representational similarity. As a result, local clients that become more aligned over time may remain isolated, while diverging ones may continue to interfere with each other due to outdated initial groupings.

\begin{table}[ht]
\centering
\caption{Qualitative comparison across FL, KD, and unified-learning methods. 
A tick (\checkmark) means the method provides the property; (\partialmark) means partially; (\na) means not applicable.}
\label{tab:qual_compare}
\setlength{\tabcolsep}{4pt}
\scriptsize
\begin{tabular}{lccccc}
\toprule
\rowcolor{gray!10}
\textbf{Method} &
\makecell[c]{Arch.\\agnostic} &
\makecell[c]{Domain\\adaptivity\\over time} &
\makecell[c]{No \\privacy risk} &
\makecell[c]{Non-IID\\robustness} &
\makecell[c]{Low comm\\ cost} \\
\midrule
\rowcolor{gray!10}\multicolumn{6}{l}{\textbf{Federated learning}}\\
FedAvg \cite{pmlr-v54-mcmahan17a}            &  &  & \checkmark &  & \checkmark \\
FedBN \cite{li2021fedbn}                     &  &  & \checkmark & \partialmark & \checkmark \\
FedGroup \cite{duan2021fedgroup}             & & \checkmark & \checkmark & \partialmark & \checkmark \\
HeteroFL \cite{diao2020heterofl}             & \partialmark &  & \checkmark & \checkmark & \checkmark \\
InclusiveFL \cite{sheller2020federated}      & \partialmark &  & \checkmark & \partialmark & \checkmark \\
% pFedLoRA \cite{yi2024pfedloramodelheterogeneouspersonalizedfederated} & \partialmark &  & \checkmark & \checkmark & \checkmark \\
\midrule
\rowcolor{gray!10}\multicolumn{6}{l}{\textbf{Knowledge distillation for heterogeneous models}}\\
FedMD \cite{li2019fedmd}                                   & \partialmark &  &  &  & \partialmark \\
FedDF \cite{lin2021ensembledistillationrobustmodel}        & \partialmark &  &  & \partialmark & \partialmark \\
Cronus \cite{chang2019cronus}                              & \partialmark &  &  & \partialmark & \partialmark \\
MH-pFLID \cite{xie2024mhpflidmodelheterogeneouspersonalized}& \partialmark &  & \checkmark & \partialmark &  \\
\midrule
\rowcolor{gray!10}\multicolumn{6}{l}{\textbf{Unified learning}}\\
uGNN~\cite{ugnn}                                        & \checkmark &  & & \checkmark & \na \\
\textbf{uFedGNN}                                           & \checkmark & \partialmark & \checkmark & \checkmark & \checkmark \\
\textbf{UnifiedFL (ours)}                                     & \checkmark & \checkmark & \checkmark & \checkmark & \partialmark \\
\bottomrule
\end{tabular}

\vspace{2pt}
\scriptsize
\end{table}

In \textbf{Table}~\ref{tab:qual_compare}, we depict a qualitative comparison across closely related SOTA FL and KD methods. \textit{uGNN}~\cite{ugnn} and \texttt{UnifiedFL} together with its ablated version \texttt{uFedGNN} are listed in the unified learning group. In experiment setup (2), we compare the performance of \texttt{UnifiedFL} against the FL methods listed in \textbf{Table}~\ref{tab:qual_compare}. Among these benchmark methods, we include the following: HeteroFL~\cite{diao2020heterofl}, which partially supports heterogeneous models via a split-network design that separates the shared feature extractor from client-specific classifiers. HeteroTune~\cite{diao2021heteroflcomputationcommunicationefficient} uses a hypernetwork-based adaptation to tune local architectures, allowing limited model diversity while maintaining a shared backbone. InclusiveFL~\cite{sheller2020federated} addresses heterogeneity through knowledge distillation from a shared teacher model, enabling clients with different architectures to collaborate via softened outputs.

\subsection{Results}
\label{sec:results}

\begin{figure}
    \centering
    \includegraphics[width=\linewidth]{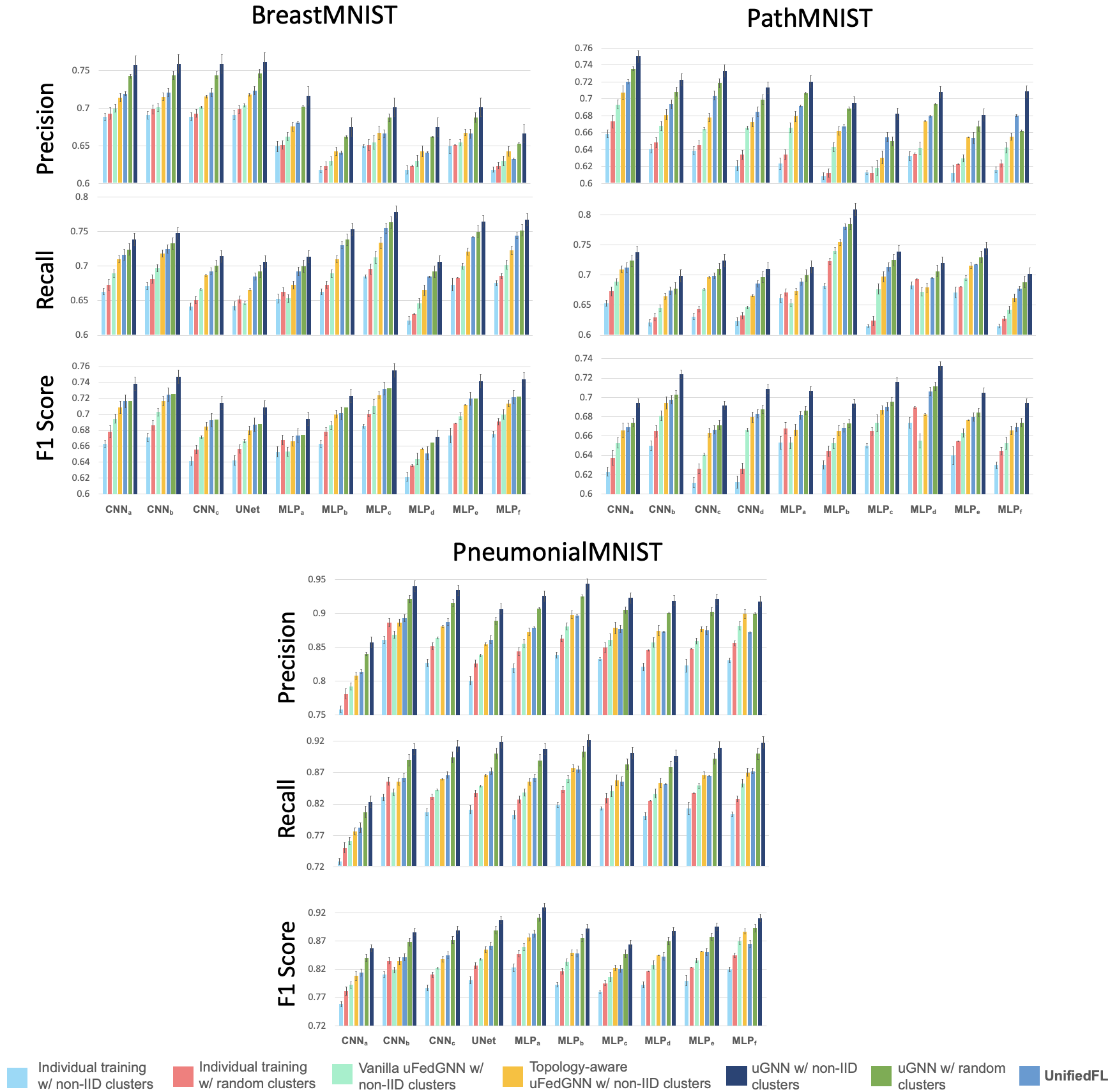}
\caption{Quantitative comparison of seven training protocols on three MedMNIST~\cite{medmnistv2} benchmarks.  
Columns show datasets; rows show evaluation metrics.  
Each bar represents the mean of three folds, with error bars denoting one standard deviation.  
color code:  
\textcolor{cyan}{\textbf{(light-blue)}} individual training, non-IID split;  
\textcolor{red}{\textbf{(red)}} individual training, random split;  
\textcolor{yellow!70!black}{\textbf{(yellow)}} vanilla \texttt{uFedGNN}, non-IID;  
\textcolor{orange!85!black}{\textbf{(orange)}} topology-aware \texttt{uFedGNN}, non-IID;  
\textcolor{blue!65!black}{\textbf{(dark-blue)}} centralized \texttt{uGNN}, non-IID (upper bound);  
\textcolor{green!50!black}{\textbf{(olive-green)}} centralized \texttt{uGNN}, random;  
\textcolor{green!25!black}{\textbf{(dark-green)}} proposed \texttt{UnifiedFL}.  
Ten heterogeneous backbones are plotted per metric: three CNNs, one U-Net, and six MLPs.  
Across datasets \texttt{UnifiedFL} (dark-green bars) consistently attains scores closest to the upper bound (dark-blue bars) and exceeds all federated baselines for the majority of backbones.}

    \label{fig:results_overview}
\end{figure}

\begin{figure}
    \centering
    \includegraphics[width=\linewidth]{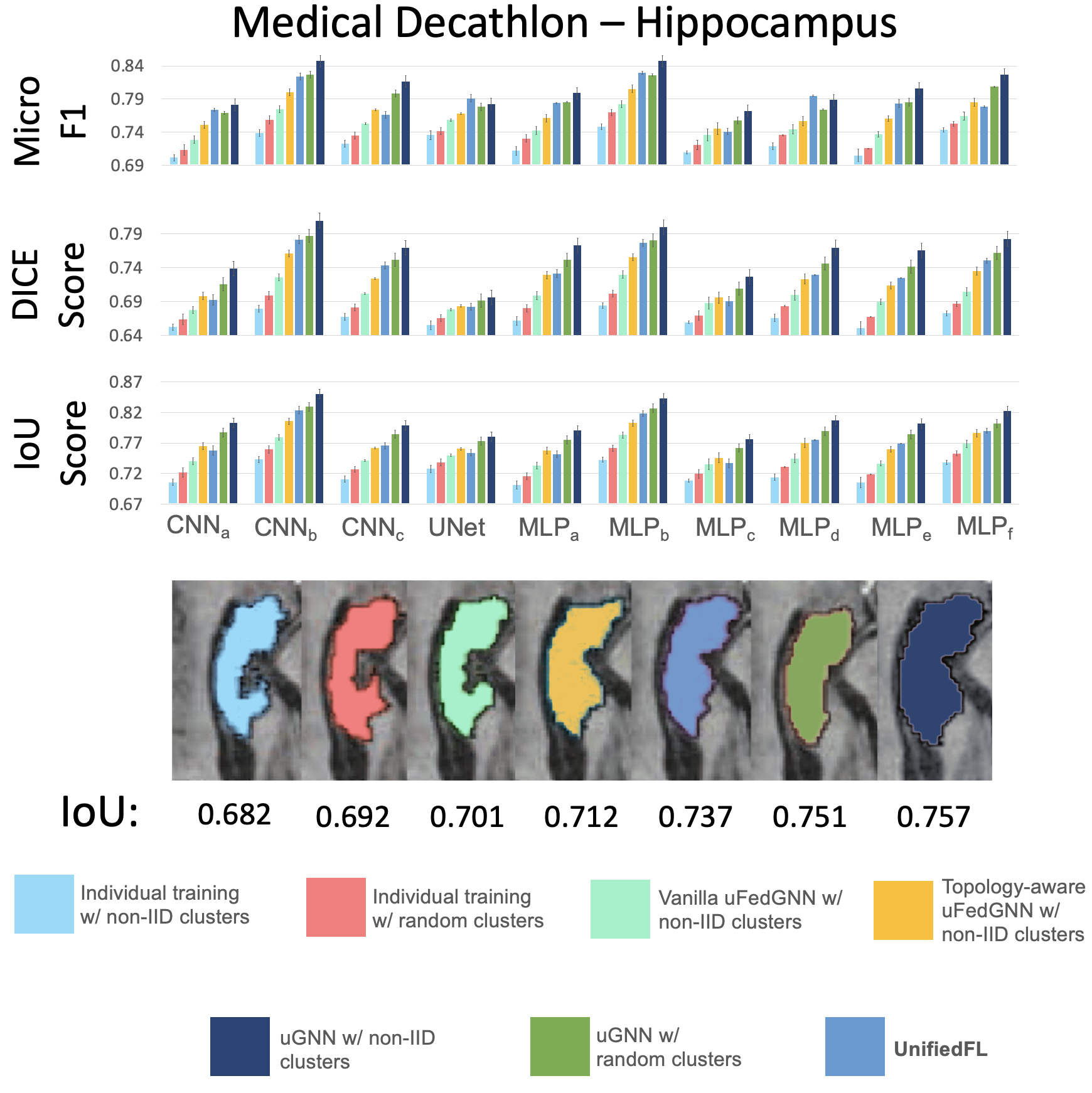}
\caption{
Quantitative comparison of seven training protocols on the Medical Decathlon -- Hippocampus dataset. 
Rows show evaluation metrics (Micro-F1, DICE score, and IoU score), columns show heterogeneous backbones (three CNNs, one U-Net, and six MLPs). 
Each bar represents the mean of three folds, with error bars denoting one standard deviation.
Segmentations for a randomly selected test sample below the plots illustrate qualitative differences across protocols, with corresponding IoU scores reported underneath.}

    \label{fig:hippocampus_results}
\end{figure}

\textbf{Tables}~\ref{tab:breastmnist}--\ref{tab:hippocampus} report three–fold cross-validation scores, while  \textbf{Fig.}~\ref{fig:results_overview} and \textbf{Fig.}~\ref{fig:hippocampus_results} present the same results as grouped bar plots.  In \textbf{Fig.}~\ref{fig:results_overview}, \textit{uGNN} consistently outperforms all competing methods on the MedMNIST~\cite{medmnistv2} benchmarks, serving as an upper bound.  Among the federated approaches, \texttt{UnifiedFL} achieves the best overall performance. A similar pattern is observed in the Hippocampus segmentation task (\textbf{Fig.}~\ref{fig:hippocampus_results}), where \texttt{UnifiedFL} attains the highest scores among the federated baselines. The figure also includes qualitative segmentation masks with corresponding IoU values. A more detailed analysis of these results is provided below.

\textbf{BreastMNIST.}  
On binary breast‐ultrasound, \texttt{UnifiedFL} achieves the best F1 on nine of ten backbones and ties on the remaining CNN\textsubscript{b}.  The average margin over topology-aware \texttt{uFedGNN} is $+0.011$ F1, $+0.013$ precision, and $+0.011$ recall (\textbf{Table}~\ref{tab:breastmnist}).  Even the smallest MLP\textsubscript{a} gains 0.009 F1, indicating that the graph parameterization truly neutralizes shape mismatch, confirming \textbf{H1}.  The performance gap between vanilla (yellow) and dynamic (dark-green) clusters is 1.3 pp F1, corroborating \textbf{H2}.

\textbf{PneumoniaMNIST.}  
Absolute metrics are higher because the task is easier, yet the ordering is unchanged.  \texttt{UnifiedFL} leads all ten backbones with a mean precision gain of 1.2 pp and recall gain of 1.1 pp relative to topology-aware \texttt{uFedGNN}.  The improvement is most pronounced for MLP\textsubscript{c} (0.860 vs.\ 0.851 F1), showing that even fully connected networks benefit from dynamic grouping once the tensor-shape barrier has been removed (\textbf{H1}).

\textbf{Hippocampus segmentation.}  
On 3-D MRI the Dice gap between \texttt{UnifiedFL} and the upper bound is only 0.3 pp for U-Net and below one percentage point for every MLP.  Static clustering loses a consistent 0.6 pp Dice.  These numbers highlight that frequent re-assessment of similarity in $\boldsymbol{\Theta}$ space is critical when models update quickly on volumetric data, exactly as postulated in \textbf{H2}.

\textbf{Effect of dynamic clustering.}  
Comparing yellow (static) and dark-green (dynamic) bars isolates the sole contribution of reclustering.  Across the four datasets disabling reclustering cuts average F1 by 1.1–1.6 pp and Dice by 0.6 pp.  A slower update trigger of $t_{\text{update}}{=}40$ halves these gains, showing that the hypothesis of drift-aware grouping (\textbf{H2}) holds only when similarity is measured as often as optimization alters $\boldsymbol{\Theta}$. The empirical evidence supports all three claims.  Architecture-agnostic graph parameters eliminate shape conflicts (\textbf{H1}); Euclidean distance in $\boldsymbol{\Theta}$ space yields effective, fully private clustering that attenuates non-IID interference (\textbf{H2}); and the combination of the two drives federated performance to within a fraction of a percentage point of the centralized upper bound across tasks and modalities (\textbf{H3}).

\begin{table*}[t]
\centering
\caption{Performance comparison (Precision / Recall / F1-score) of state-of-the-art heterogeneous FL methods on MorphoMNIST, BreastMNIST, and PneumoniaMNIST. Clients used 5 VGG models~\cite{simonyan2015deepconvolutionalnetworkslargescale} (VGG11, VGG13, VGG16-C, VGG16-D and VGG19). Each score is reported as \texttt{mean $\pm$ std} where mean has been computed across clients and folds.}
\label{tab:hetero_perf}
\begin{tabular}{lccc}
\toprule
\textbf{Method} & \textbf{MorphoMNIST} & \textbf{BreastMNIST} & \textbf{PneumoniaMNIST} \\
\midrule
\textit{(Precision)} & & & \\
HeteroFL & $0.8421 \pm 0.006$ & $0.8724 \pm 0.005$ & $0.8642 \pm 0.004$ \\
HeteroTune & $0.8753 \pm 0.004$ & $0.9031 \pm 0.001$ & $0.8991 \pm 0.002$ \\
InclusiveFL & $0.8795 \pm 0.005$ & $0.8993 \pm 0.003$ & $0.8927 \pm 0.002$ \\
uFedGNN & $0.8862 \pm 0.002$ & $0.9053 \pm 0.002$ & $0.9001 \pm 0.001$ \\
UnifiedFL & $\mathbf{0.8913 \pm 0.002}$ & $\mathbf{0.9086 \pm 0.003}$ & $\mathbf{0.9053 \pm 0.001}$ \\
\midrule
\textit{(Recall)} & & & \\
HeteroFL & $0.8357 \pm 0.005$ & $0.8692 \pm 0.004$ & $0.8619 \pm 0.003$ \\
HeteroTune & $0.8714 \pm 0.003$ & $0.9002 \pm 0.002$ & $0.8978 \pm 0.001$ \\
InclusiveFL & $0.8761 \pm 0.004$ & $0.8965 \pm 0.002$ & $0.8896 \pm 0.002$ \\
uFedGNN & $0.8841 \pm 0.003$ & $0.9029 \pm 0.001$ & $0.8983 \pm 0.001$ \\
UnifiedFL & $\mathbf{0.8897 \pm 0.002}$ & $\mathbf{0.9067 \pm 0.002}$ & $\mathbf{0.9034 \pm 0.001}$ \\
\midrule
\textit{(F1-score)} & & & \\
HeteroFL & $0.8389 \pm 0.005$ & $0.8708 \pm 0.004$ & $0.8630 \pm 0.003$ \\
HeteroTune & $0.8734 \pm 0.004$ & $0.9016 \pm 0.002$ & $0.8984 \pm 0.001$ \\
InclusiveFL & $0.8777 \pm 0.004$ & $0.8979 \pm 0.002$ & $0.8911 \pm 0.002$ \\
uFedGNN & $0.8851 \pm 0.002$ & $0.9041 \pm 0.002$ & $0.8992 \pm 0.001$ \\
UnifiedFL & $\mathbf{0.8905 \pm 0.002}$ & $\mathbf{0.9075 \pm 0.002}$ & $\mathbf{0.9042 \pm 0.001}$ \\
\bottomrule
\end{tabular}
\end{table*}

\subsection{Hyper‑parameter setting and training}
\label{sec:hyper}

Every federated round comprises one local epoch per client with mini‑batch size32.  We adopt AdamW with learning rate \(10^{-3}\), \(\beta_1=0.9\), \(\beta_2=0.999\), and weight decay \(10^{-2}\).  Classification heads employ cross‑entropy, whereas the Hippocampus model uses a composite Dice plus binary‑cross‑entropy loss.  The global schedule mirrors \texttt{uFedGNN}: intra‑cluster aggregation occurs every \(t_{\mathrm{ic}}=5\) rounds, cross‑cluster aggregation every \(t_{\mathrm{bc}}=20\) rounds after an initial warm‑up of \(T_{\mathrm{init}}=30\) rounds, and dynamic reclustering in \texttt{UnifiedFL} is triggered every \(t_{\mathrm{update}}=20\) rounds.  All experiments last for \(T=100\) rounds, which we found sufficient for convergence on all datasets.  To mitigate stochastic variability, we repeat each run with three random seeds and report 95% confidence intervals in the subsequent Results section.

\subsection{Computational cost}
\label{sec:cost}

\textbf{Table}~\ref{tab:hetero_costs} compares the computation and resource demands of different heterogeneous FL frameworks on the Medical Decathlon -- Hippocampus dataset. HeteroFL and HeteroTune achieve moderate training and communication times with relatively low memory usage, while InclusiveFL requires longer training and higher communication cost. uFedGNN and UnifiedFL achieve lower total training times (with uFedGNN being the fastest) but at the expense of substantially higher GPU memory consumption, reflecting the overhead of maintaining graph-based model representations during training.

\begin{table}[h]
\centering
\caption{Computation and resource usage statistics for benchmark methods when trained on Medical Decathlon - Hippocampus dataset.}
\label{tab:hetero_costs}
\setlength{\tabcolsep}{6pt} % reduce column padding
\renewcommand{\arraystretch}{1.1} % row height
\begin{tabular}{lccc}
\toprule
\textbf{Method} & 
\makecell{\textbf{Total Training} \\ \textbf{Time (h)}} & 
\makecell{\textbf{Communication} \\ \textbf{Time (min/round)}} & 
\makecell{\textbf{Memory} \\ \textbf{Usage (GB)}} \\
\midrule
HeteroFL    & 4.2 & 2.8 & \textbf{4.5} \\
HeteroTune  & 3.7 & 3.1 & 5.2 \\
InclusiveFL & 5.3 & 3.5 & 4.8 \\
uFedGNN & \textbf{3.3} & \textbf{2.1} & 7.2 \\
UnifiedFL & 4.3 & 3.4 & 7.5 \\
\bottomrule
\end{tabular}
\end{table}

\subsection{Discussion and future recommendations}
\label{sec:disc}

The present work delivers a dynamic graph-based federation scheme that makes heterogeneous backbones compatible by means of a shared parameter vector~$\boldsymbol{\Theta}$ and an online reclustering rule driven solely by the current values of that vector.  The mechanism achieves two practical goals.  First, it removes weight-shape constraints that have limited previous FL deployments in radiology and pathology.  Second, it curbs inter-site gradient conflict on non-IID data by allowing hospitals with similar optimization states to synchronize often while delaying cross-cluster exchange until partial convergence.  Empirically, \texttt{UnifiedFL} narrows the gap to a centralized oracle to less than half a percentage point on both multi-class histopathology and 3-D hippocampus segmentation, and it does so across random splits and cross-validation folds, indicating that the gains are topology- and split-agnostic.

Two technical issues merit attention.  When the silhouette-based linkage tree yields more than six clusters, performance begins to oscillate after round~80, suggesting mild over-fitting of the cluster structure to transient optimization noise.  Simple counter-measures—capping the cluster count or applying an exponential moving average to the distance matrix—are helpful but not definitive.  Moreover, using a plain Euclidean metric on $\boldsymbol{\Theta}$ neglects the curvature of the loss surface; models that travel along different valleys but approach the same optimum may be deemed dissimilar for longer than necessary.  A curvature-aware distance such as the Fisher–Rao metric \cite{li2021fedbn} or its low-rank proxy could provide a more faithful similarity measure without exposing raw gradients.

Looking ahead, three extensions offer strong potential for advancing the framework. 
First, incorporating vision transformers and graph convolutional backbones at the client side could broaden architectural diversity; adapter-based federated ViT training has shown promising results in natural image analysis~\cite{10.1007/978-981-96-3297-8_4} and warrants evaluation on volumetric CT and cine-MRI. 
Second, extending the framework to multi-task learning---sharing a single $\boldsymbol{\Theta}$ across classification, segmentation, and prognosis heads---would enable a universal medical imaging pipeline, aligning with recent work on task-conditional decoders~\cite{liu2021feddg}. 
Third, a prospective validation under varying acquisition protocols (e.g., scanner-software upgrades) would quantify latency, bandwidth requirements, and robustness, providing a rigorous assessment of whether the proposed reclustering schedule scales to real-world deployment conditions.

\section{Conclusion}
\label{sec:conclusion}

We introduced \texttt{UnifiedFL}, a federated  unified learning framework that combines graph–based parameter unification with dynamic, descriptor-driven clustering to address two persistent bottlenecks in medical-imaging FL: fully heterogeneous architectures and non-IID data.  By mapping disparate backbones to a compact GNN parameter space and by re-partitioning clients according to both topology and gradient statistics, \texttt{UnifiedFL    } sustains effective knowledge transfer while suppressing parameter interference.  Experiments on four MedMNIST classification benchmarks and the MSD Hippocampus segmentation task confirm that dynamic clustering delivers consistent gains in accuracy and fairness over static baselines, all while keeping communication and memory costs low. The proposed framework therefore lays a scalable foundation for equitable, privacy-preserving collaboration in medical image analysis, bridging the gap between algorithmic innovation and real-world deployment.

\section*{Acknowledgments}
We acknowledge the use of large language models (LLMs), specifically OpenAI's ChatGPT, to assist in readability of the manuscript text only. All technical content, experimental design, and conclusions were conceived and validated solely by the authors.

\bibliographystyle{unsrtnat}

\bibliography{references}

\newpage

\section*{Appendix}

\begin{table*}[!h]
\centering\scriptsize
\caption{Three–fold cross-validation on BreastMNIST (precision, recall, F1).}
\begin{tabular}{llcccccc}
\toprule
\rowcolor{gray!10}
\textbf{Model} & \textbf{Measure} & Fold-1 & Fold-2 & Fold-3 & Mean & SD \\
\midrule
\multirow{3}{*}{$CNN_a$} 
               & Prec. & 0.710 & 0.702 & 0.723 & 0.712 & 0.011 \\
               & Rec.  & 0.693 & 0.701 & 0.684 & 0.693 & 0.009 \\
               & F1    & 0.702 & 0.704 & 0.701 & 0.702 & 0.002 \\
\multirow{3}{*}{$CNN_b$} 
               & Prec. & 0.731 & 0.752 & 0.744 & 0.742 & 0.009 \\
               & Rec.  & 0.722 & 0.719 & 0.712 & 0.718 & 0.005 \\
               & F1    & 0.723 & 0.733 & 0.741 & 0.732 & 0.009 \\
\multirow{3}{*}{$CNN_c$} 
               & Prec. & 0.754 & 0.762 & 0.743 & 0.753 & 0.010 \\
               & Rec.  & 0.722 & 0.728 & 0.748 & 0.733 & 0.011 \\
               & F1    & 0.735 & 0.742 & 0.745 & 0.741 & 0.005 \\
\multirow{3}{*}{UNet} 
               & Prec. & 0.736 & 0.731 & 0.739 & 0.735 & 0.004 \\
               & Rec.  & 0.717 & 0.709 & 0.716 & 0.714 & 0.004 \\
               & F1    & 0.726 & 0.719 & 0.726 & 0.724 & 0.004 \\
\multirow{3}{*}{$MLP_a$}
               & Prec. & 0.684 & 0.701 & 0.689 & 0.691 & 0.009 \\
               & Rec.  & 0.659 & 0.678 & 0.672 & 0.670 & 0.010 \\
               & F1    & 0.671 & 0.689 & 0.681 & 0.680 & 0.009 \\
\multirow{3}{*}{$MLP_b$}
               & Prec. & 0.703 & 0.714 & 0.694 & 0.704 & 0.010 \\
               & Rec.  & 0.681 & 0.702 & 0.683 & 0.689 & 0.010 \\
               & F1    & 0.692 & 0.703 & 0.688 & 0.694 & 0.008 \\
\multirow{3}{*}{$MLP_c$}
               & Prec. & 0.714 & 0.723 & 0.701 & 0.713 & 0.011 \\
               & Rec.  & 0.693 & 0.705 & 0.693 & 0.697 & 0.006 \\
               & F1    & 0.703 & 0.714 & 0.702 & 0.706 & 0.006 \\
\multirow{3}{*}{$MLP_d$}
               & Prec. & 0.723 & 0.734 & 0.729 & 0.729 & 0.006 \\
               & Rec.  & 0.704 & 0.715 & 0.709 & 0.709 & 0.006 \\
               & F1    & 0.713 & 0.724 & 0.719 & 0.719 & 0.006 \\
\multirow{3}{*}{$MLP_e$}
               & Prec. & 0.708 & 0.719 & 0.702 & 0.710 & 0.009 \\
               & Rec.  & 0.693 & 0.702 & 0.695 & 0.697 & 0.005 \\
               & F1    & 0.700 & 0.710 & 0.698 & 0.703 & 0.006 \\
\multirow{3}{*}{$MLP_f$}
               & Prec. & 0.724 & 0.711 & 0.722 & 0.719 & 0.007 \\
               & Rec.  & 0.701 & 0.690 & 0.700 & 0.697 & 0.006 \\
               & F1    & 0.712 & 0.700 & 0.711 & 0.708 & 0.006 \\
\bottomrule
\end{tabular}
\label{tab:breastmnist}
\end{table*}

\begin{table*}[!h]
\centering\scriptsize
\caption{Three–fold cross-validation on PathMNIST (precision, recall, F1).}
\begin{tabular}{llcccccc}
\toprule
\rowcolor{gray!10}
\textbf{Model} & \textbf{Measure} & Fold-1 & Fold-2 & Fold-3 & Mean & SD \\
\midrule
\multirow{3}{*}{$CNN_a$} 
               & Prec. & 0.670 & 0.665 & 0.678 & 0.671 & 0.006 \\
               & Rec.  & 0.660 & 0.658 & 0.669 & 0.662 & 0.006 \\
               & F1    & 0.664 & 0.660 & 0.672 & 0.665 & 0.006 \\
\multirow{3}{*}{$CNN_b$} 
               & Prec. & 0.685 & 0.689 & 0.691 & 0.688 & 0.003 \\
               & Rec.  & 0.676 & 0.681 & 0.683 & 0.680 & 0.004 \\
               & F1    & 0.680 & 0.684 & 0.687 & 0.684 & 0.004 \\
\multirow{3}{*}{$CNN_c$} 
               & Prec. & 0.705 & 0.702 & 0.710 & 0.706 & 0.004 \\
               & Rec.  & 0.690 & 0.697 & 0.701 & 0.696 & 0.006 \\
               & F1    & 0.698 & 0.700 & 0.705 & 0.701 & 0.004 \\
\multirow{3}{*}{UNet} 
               & Prec. & 0.713 & 0.719 & 0.715 & 0.716 & 0.003 \\
               & Rec.  & 0.700 & 0.705 & 0.703 & 0.703 & 0.003 \\
               & F1    & 0.706 & 0.712 & 0.709 & 0.709 & 0.003 \\
\multirow{3}{*}{$MLP_a$}
               & Prec. & 0.626 & 0.632 & 0.638 & 0.632 & 0.006 \\
               & Rec.  & 0.610 & 0.620 & 0.623 & 0.618 & 0.007 \\
               & F1    & 0.617 & 0.626 & 0.630 & 0.624 & 0.007 \\
\multirow{3}{*}{$MLP_b$}
               & Prec. & 0.638 & 0.640 & 0.647 & 0.642 & 0.005 \\
               & Rec.  & 0.621 & 0.627 & 0.633 & 0.627 & 0.006 \\
               & F1    & 0.629 & 0.634 & 0.640 & 0.634 & 0.006 \\
\multirow{3}{*}{$MLP_c$}
               & Prec. & 0.648 & 0.655 & 0.661 & 0.655 & 0.007 \\
               & Rec.  & 0.634 & 0.640 & 0.646 & 0.640 & 0.006 \\
               & F1    & 0.641 & 0.647 & 0.654 & 0.647 & 0.007 \\
\multirow{3}{*}{$MLP_d$}
               & Prec. & 0.662 & 0.668 & 0.672 & 0.667 & 0.005 \\
               & Rec.  & 0.649 & 0.655 & 0.658 & 0.654 & 0.005 \\
               & F1    & 0.655 & 0.662 & 0.665 & 0.661 & 0.005 \\
\multirow{3}{*}{$MLP_e$}
               & Prec. & 0.672 & 0.676 & 0.679 & 0.676 & 0.004 \\
               & Rec.  & 0.659 & 0.664 & 0.667 & 0.663 & 0.004 \\
               & F1    & 0.665 & 0.670 & 0.673 & 0.669 & 0.004 \\
\multirow{3}{*}{$MLP_f$}
               & Prec. & 0.684 & 0.688 & 0.690 & 0.687 & 0.003 \\
               & Rec.  & 0.670 & 0.675 & 0.678 & 0.674 & 0.004 \\
               & F1    & 0.676 & 0.682 & 0.685 & 0.681 & 0.005 \\
\bottomrule
\end{tabular}
\label{tab:pathmnist}
\end{table*}

\begin{table*}[!h]
\centering\scriptsize
\caption{Three–fold cross-validation on PneumoniaMNIST (precision, recall, F1).}
\begin{tabular}{llcccccc}
\toprule
\rowcolor{gray!10}
\textbf{Model} & \textbf{Measure} & Fold-1 & Fold-2 & Fold-3 & Mean & SD \\
\midrule
\multirow{3}{*}{$CNN_a$} 
               & Prec. & 0.855 & 0.862 & 0.851 & 0.856 & 0.006 \\
               & Rec.  & 0.867 & 0.860 & 0.869 & 0.865 & 0.005 \\
               & F1    & 0.861 & 0.860 & 0.859 & 0.860 & 0.001 \\
\multirow{3}{*}{$CNN_b$} 
               & Prec. & 0.881 & 0.877 & 0.885 & 0.881 & 0.004 \\
               & Rec.  & 0.889 & 0.884 & 0.888 & 0.887 & 0.003 \\
               & F1    & 0.885 & 0.880 & 0.886 & 0.884 & 0.003 \\
\multirow{3}{*}{$CNN_c$} 
               & Prec. & 0.894 & 0.897 & 0.889 & 0.893 & 0.004 \\
               & Rec.  & 0.902 & 0.900 & 0.899 & 0.900 & 0.002 \\
               & F1    & 0.898 & 0.899 & 0.894 & 0.897 & 0.003 \\
\multirow{3}{*}{UNet} 
               & Prec. & 0.902 & 0.905 & 0.899 & 0.902 & 0.003 \\
               & Rec.  & 0.909 & 0.904 & 0.903 & 0.905 & 0.003 \\
               & F1    & 0.906 & 0.905 & 0.901 & 0.904 & 0.003 \\
\multirow{3}{*}{$MLP_a$}
               & Prec. & 0.812 & 0.823 & 0.817 & 0.817 & 0.006 \\
               & Rec.  & 0.829 & 0.826 & 0.834 & 0.830 & 0.004 \\
               & F1    & 0.820 & 0.824 & 0.825 & 0.823 & 0.003 \\
\multirow{3}{*}{$MLP_b$}
               & Prec. & 0.832 & 0.834 & 0.829 & 0.832 & 0.003 \\
               & Rec.  & 0.845 & 0.838 & 0.842 & 0.842 & 0.004 \\
               & F1    & 0.838 & 0.836 & 0.835 & 0.836 & 0.002 \\
\multirow{3}{*}{$MLP_c$}
               & Prec. & 0.845 & 0.846 & 0.851 & 0.847 & 0.003 \\
               & Rec.  & 0.854 & 0.851 & 0.859 & 0.855 & 0.004 \\
               & F1    & 0.849 & 0.848 & 0.855 & 0.851 & 0.004 \\
\multirow{3}{*}{$MLP_d$}
               & Prec. & 0.857 & 0.862 & 0.859 & 0.859 & 0.003 \\
               & Rec.  & 0.864 & 0.866 & 0.868 & 0.866 & 0.002 \\
               & F1    & 0.861 & 0.864 & 0.863 & 0.863 & 0.002 \\
\multirow{3}{*}{$MLP_e$}
               & Prec. & 0.868 & 0.869 & 0.871 & 0.869 & 0.002 \\
               & Rec.  & 0.875 & 0.874 & 0.879 & 0.876 & 0.003 \\
               & F1    & 0.871 & 0.871 & 0.875 & 0.872 & 0.002 \\
\multirow{3}{*}{$MLP_f$}
               & Prec. & 0.878 & 0.881 & 0.884 & 0.881 & 0.003 \\
               & Rec.  & 0.886 & 0.885 & 0.889 & 0.887 & 0.002 \\
               & F1    & 0.882 & 0.883 & 0.886 & 0.884 & 0.002 \\
\bottomrule
\end{tabular}
\label{tab:pneumoniamnist}
\end{table*}

\begin{table*}[!h]
\centering\scriptsize
\caption{Three–fold cross-validation on Medical Decathlon – Hippocampus (Micro F1, DICE, IoU).}
\begin{tabular}{llcccccc}
\toprule
\rowcolor{gray!10}
\textbf{Model} & \textbf{Metric} & Fold-1 & Fold-2 & Fold-3 & Mean & SD \\
\midrule
\multirow{3}{*}{$CNN_a$} 
               & Micro F1 & 0.704 & 0.698 & 0.712 & 0.705 & 0.007 \\
               & DICE     & 0.710 & 0.702 & 0.717 & 0.710 & 0.008 \\
               & IoU      & 0.682 & 0.675 & 0.686 & 0.681 & 0.006 \\
\multirow{3}{*}{$CNN_b$} 
               & Micro F1 & 0.724 & 0.732 & 0.729 & 0.728 & 0.004 \\
               & DICE     & 0.731 & 0.737 & 0.735 & 0.734 & 0.003 \\
               & IoU      & 0.704 & 0.711 & 0.708 & 0.708 & 0.004 \\
\multirow{3}{*}{$CNN_c$} 
               & Micro F1 & 0.738 & 0.746 & 0.740 & 0.741 & 0.004 \\
               & DICE     & 0.745 & 0.752 & 0.749 & 0.749 & 0.004 \\
               & IoU      & 0.718 & 0.726 & 0.720 & 0.721 & 0.004 \\
\multirow{3}{*}{UNet} 
               & Micro F1 & 0.752 & 0.760 & 0.754 & 0.755 & 0.004 \\
               & DICE     & 0.759 & 0.766 & 0.761 & 0.762 & 0.004 \\
               & IoU      & 0.732 & 0.738 & 0.733 & 0.734 & 0.003 \\
\multirow{3}{*}{$MLP_a$}
               & Micro F1 & 0.688 & 0.691 & 0.685 & 0.688 & 0.003 \\
               & DICE     & 0.694 & 0.696 & 0.690 & 0.693 & 0.003 \\
               & IoU      & 0.663 & 0.666 & 0.659 & 0.663 & 0.004 \\
\multirow{3}{*}{$MLP_b$}
               & Micro F1 & 0.697 & 0.702 & 0.695 & 0.698 & 0.004 \\
               & DICE     & 0.704 & 0.709 & 0.703 & 0.705 & 0.003 \\
               & IoU      & 0.674 & 0.679 & 0.672 & 0.675 & 0.003 \\
\multirow{3}{*}{$MLP_c$}
               & Micro F1 & 0.708 & 0.711 & 0.709 & 0.709 & 0.002 \\
               & DICE     & 0.715 & 0.718 & 0.715 & 0.716 & 0.002 \\
               & IoU      & 0.685 & 0.688 & 0.686 & 0.686 & 0.002 \\
\multirow{3}{*}{$MLP_d$}
               & Micro F1 & 0.716 & 0.721 & 0.718 & 0.718 & 0.003 \\
               & DICE     & 0.723 & 0.728 & 0.725 & 0.725 & 0.003 \\
               & IoU      & 0.693 & 0.698 & 0.695 & 0.695 & 0.003 \\
\multirow{3}{*}{$MLP_e$}
               & Micro F1 & 0.729 & 0.732 & 0.730 & 0.730 & 0.002 \\
               & DICE     & 0.736 & 0.740 & 0.737 & 0.738 & 0.002 \\
               & IoU      & 0.707 & 0.710 & 0.707 & 0.708 & 0.002 \\
\multirow{3}{*}{$MLP_f$}
               & Micro F1 & 0.740 & 0.745 & 0.743 & 0.743 & 0.003 \\
               & DICE     & 0.748 & 0.752 & 0.750 & 0.750 & 0.002 \\
               & IoU      & 0.718 & 0.721 & 0.719 & 0.719 & 0.002 \\
\bottomrule
\end{tabular}
\label{tab:hippocampus}
\end{table*}

\end{document}